\documentclass{article}
\usepackage{amssymb}

\usepackage[preprint]{corl_2025} 
\usepackage{amsmath}
\usepackage[makeroom]{cancel}
\usepackage{booktabs}
\usepackage{multirow}

\usepackage{makecell} 
\usepackage{graphicx}      
\usepackage{wrapfig}
\usepackage{subfig}
\usepackage{enumitem}

\usepackage[table, dvipsnames]{xcolor} 
\usepackage{booktabs}      
\usepackage{multirow}      

\title{MirrorDuo: Reflection-Consistent Visuomotor Learning from Mirrored Demonstration Pairs}

\author{
Zheyu Zhuang$^{1}$\thanks{Equal contribution}, 
Ruiyu Wang$^{1\ast}$, 
Giovanni Luca Marchetti$^{2}$, \\
\textbf{Florian T. Pokorny$^{1}$, 
Danica Kragic$^{1}$}\\
$^1$Division of Robotics, Perception and Learning, $^2$Department of Mathematics \\
KTH Royal Institute of Technology, Stockholm, Sweden \\
}

\begin{document}
\maketitle


\begin{abstract}
Image-based behaviour cloning leverages demonstrations captured from ubiquitous RGB cameras. However, it remains constrained by the cost of collecting diverse demos, especially for generalizing across workspace variations.
We propose MirrorDuo, a reflection-based formulation that operates on image, proprioception, and full 6-DoF end-effector action tuples, generating a mirrored counterpart for each original demonstration, effectively achieving ``collect one, get one for free''.
It can be applied as a data augmentation strategy for existing learning pipelines, such as standard behaviour cloning or diffusion policy, or as a structural prior for reflection-equivariant policy networks.
By leveraging the overlap between the original and mirrored domains, MirrorDuo achieves significantly improved performance under the same data budget when demonstrations are evenly distributed across both sides of the workspace.
When demonstrations are confined to one side, MirrorDuo enables efficient skill transfer to the mirrored workspace with as few as zero or five demos in the target arrangement.
\,\href{https://github.com/zheyu-zhuang/mirror-duo}{\raisebox{-0.2ex}{\includegraphics[height=1em]{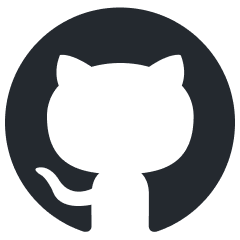}}}
\end{abstract}
\vspace{-2mm}

\section{Introduction}

\begin{wrapfigure}{r}{0.45\textwidth}
    \vspace{-9pt}
    \centering
    \includegraphics[width=0.95\linewidth]{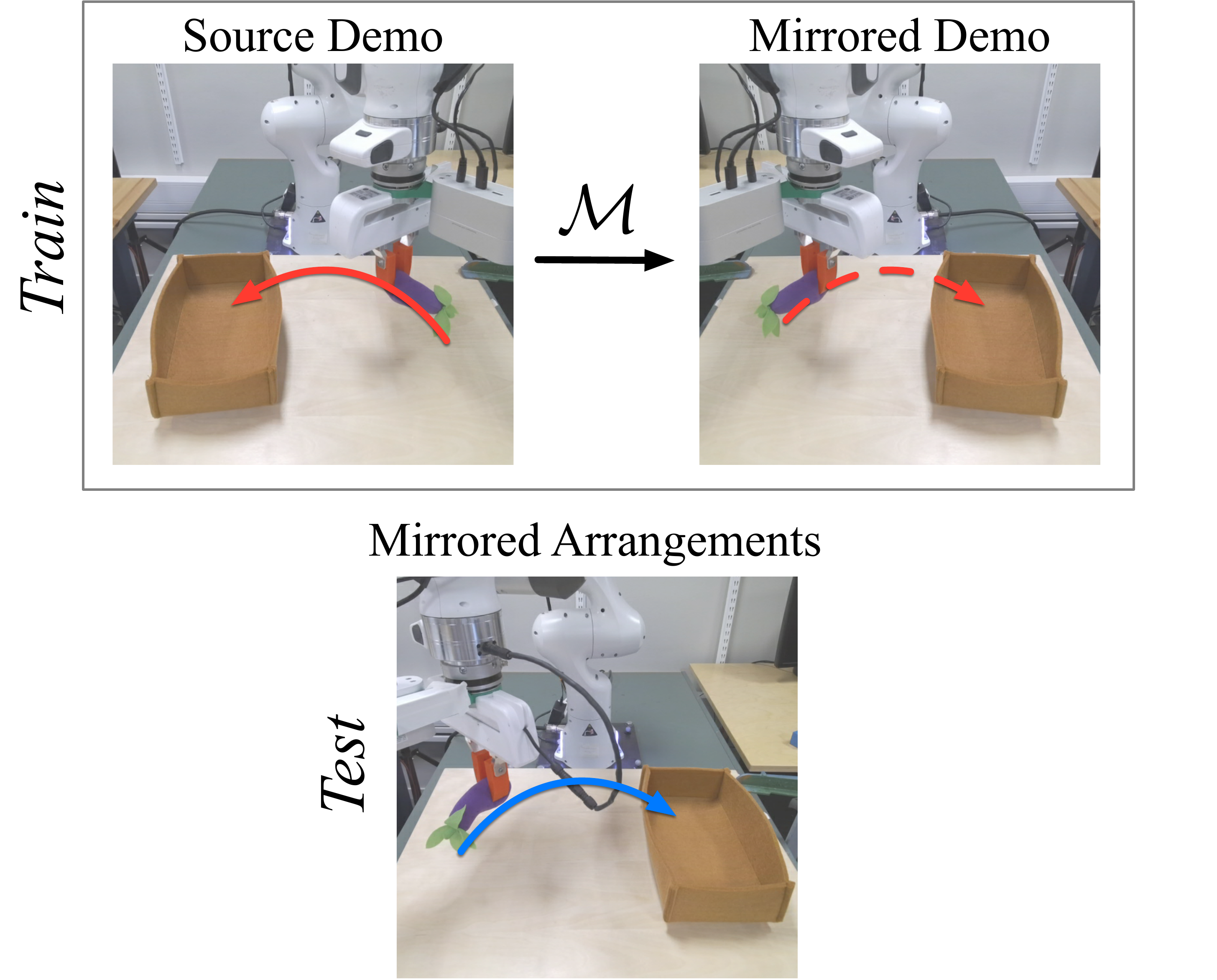}
    \caption{\textbf{Illustration of MirrorDuo ($\mathcal{M}$).} Mirroring a source demo to synthesis paired demo in the mirrored arrangement.}
    \label{fig:title_image}
    \vspace{-5mm}
\end{wrapfigure}

Behaviour Cloning (BC) from visual demonstrations holds promise for scalable skill acquisition in real-world environments. 
Still, it is constrained by the cost of collecting diverse data, particularly in settings with spatial variation of target objects or asymmetric scene layouts~\citep{gao2024,xue2025demogen}, see Fig.~\ref {fig:title_image}.

Unlike image-based BC, methods leveraging 3D inputs (e.g., point clouds) can exploit spatial roto-translation equivariance to improve data efficiency. 
For example, $\mathrm{SE}(3)$ rigid transformations on the 3D inputs preserve spatial relationships between the robot and objects, enabling policies to generalize to new configurations by synthesizing transformed demonstrations~\cite {eisnerdeep_se3_equi_placement, ryu_equivariant_se3_manipulation_learning, yang_equibot_sim3}.
In contrast to the natural $\mathrm{SE}(3)$ equivariance emerging in 3D representations, applying 3D transformations to 2D images often produces inconsistent effects, with meaningful transformations restricted to simplified settings such as planar top-down views~\cite{wang_equi_rl}.
While prior work~\cite{wang_equivariant_diff_po} approximates $\mathrm{SO}(2)$ symmetry by rotating third-person images, its gains remain limited to in-domain settings and fall short of the broader generalization supported by 3D inputs.


One underexplored source of structure for image-based policies is \textit{reflection} symmetry. 
Many manipulation tasks have mirrored variants, for example, a pick-and-place on the left can be reflected to the right, and a pushing trajectory often has a counterpart on the opposite side. 
Such image-space symmetries are typically faithful to end-effector–object relations, though may be affected by visual artifacts, as illustrated in Fig.~\ref{fig:title_image}.
Prior work~\cite{jia2023seil, wang2023_equi_latent_symm} has considered mirroring image and action pairs, but only in simplified settings (e.g., top-down views or $\mathrm{SE}(2)$ actions) and without addressing generalization to unseen mirrored scenarios.

We introduce MirrorDuo, a general formulation for incorporating reflection symmetry into generic visuomotor learning settings. 
MirrorDuo applies mirroring jointly to RGB observations (both eye-in-hand and third-person), proprioceptive inputs, and 6-DoF actions, producing semantically and physically consistent pairs. This enables two complementary use cases: (1) as a data augmentation strategy that extends coverage to mirrored configurations, and (2) as a prior for learning reflection-equivariant policies that generalize by construction. 
While mirroring introduces visual asymmetries (Fig.~\ref{fig:visual_asymm_from_robot}), policies trained on demonstrations from one side achieve near in-domain performance on the mirrored side with substantially fewer demonstrations. 
When demonstrations from both sides are available, this symmetry can be further leveraged to improve data efficiency, yielding competent performance with fewer demonstrations overall.
 
\section{Related Work}
\label{sec:related_work}
\textbf{Image-based Behaviour Cloning (BC)} learns policies that map observations to actions from demonstrations \( (\mathbf{o}_t, \mathbf{a}_t) \), where observations include multi-view images \( \{\mathbf{I}_t^{(c)}\}_{c \in \mathcal{C}} \), with \( \mathcal{C} \) denoting the set of camera views (e.g., eye-in-hand, third-person), and proprioceptive states \( \mathbf{s}_t \). Early explicit-policy methods~\citep{NIPS1988_812b4ba2, RahmatizadehABL17, robomimic2021} struggle with the multi-modal nature of human actions. Implicit approaches, such as energy-based models~\citep{florence2021implicit}, and more recent diffusion-based policies~\citep{chi2023diffusionpolicy, prasad2024consistencypolicy, ze2024_3d_diffpo}, better capture action distributions via generative modeling.

\textbf{Geometric Symmetries in Robotic Manipulation} are rooted in the structure of rigid transformations in the state space, such as complete 3D rigid transformations, i.e., $\mathrm{SE}(3)$.
When the end-effector and object are transformed by the same rigid transformation, the resulting trajectory, expressed in the relative frame, remains invariant, providing a theoretical foundation for gains in data efficiency.
On the data generation side, recent works~\cite{mandlekar2023mimicgen,xue2025demogen,hoque2024intervengen,garrett2024skillmimicgen,jiang2024dexmimicgen} exploit such symmetries to augment demonstrations by replaying transformed trajectories for transformed objects.
On the policy learning side, recent works have shown compelling results in incorporating these symmetries through equivariant policies~\cite{eisnerdeep_se3_equi_placement, ryu_equivariant_se3_manipulation_learning, yang_equibot_sim3, jia2023seil, wang2023_equi_latent_symm}. In particular,~\cite{yang_equibot_sim3, wang_equivariant_diff_po} leverage equivariant properties within the diffusion process~\cite{hoogeboom2022equivariant_diff}.
These approaches often represent observations and actions in a 3D geometric form, which are naturally suited for $\mathrm{SE(3)}$ or $\mathrm{SE(2)}$ transformations due to their invariance under such transformations. However, in image-based BC settings, the observations are 2D projections of the 3D world, while the actions and robot states remain in 3D. 
The $\mathrm{SO}(2)$-Equivariant Diffusion Policy~\cite{wang_equivariant_diff_po} uses image inputs and enforces equivariance by rotating third-person images and transforming proprioception and actions to emulate global scene rotation. 
Other approaches~\cite{jia2023seil, wang2023_equi_latent_symm} incorporate image reflection but with simplified observation and action spaces.

\section{Methodology}
\label{sec:methodology}
\textbf{Mirroring in Pose Space for Static Cameras.} To ensure consistency with image mirroring (horizontal flip), we define an analogous mirroring operation in pose space. 
Let \( {}^C\mathbf{X}_{H_t}^{\phantom{*}} \in \mathrm{SE}(3) \) denote the end-effector pose \( \mathbf{X}_{H_t}^{\phantom{*}} \) at time $t$ expressed in the camera frame \( \mathbf{X}_C \). 
Henceforth, we drop the time index for brevity. 
Under a standard pinhole camera model, the mirrored pose is given by $
    {}^C\mathbf{X}^*_{H} = \mathbf{E} \, {}^C\mathbf{X}_{H}^{\phantom{*}} \,\mathbf{E},
    \label{eq:mirror_raw_form}
$
where $\mathbf{E} = \mathrm{diag}([-1, 1, 1, 1])$. Mapping back to the world frame yields:
\begin{equation}
    \mathbf{X}^*_{H} = \mathbf{X}_C \, \mathbf{E} \, \mathbf{X}_C^{-1} \, \mathbf{X}_{H}^{\phantom{*}} \, \mathbf{E}.
    \label{eq:abs_pose_mirroring}
\end{equation}
This formulation also applies to absolute actions, with \( \mathbf{a}_t := \mathbf{X}_{H_{t+1}} \).
Gripper states and actions (e.g., open/close) are omitted, as they do not exhibit geometric structures affected by mirroring.
 
\textbf{Eliminating Dependency on Camera Extrinsics.}
While Eq.~\eqref{eq:abs_pose_mirroring} guarantees geometric consistency with image mirroring, it depends on the camera extrinsics \( \mathbf{X}_C \), which are often unavailable in BC datasets. 
To address this, we express every pose in a local coordinate system.  
We use two equivalent forms:  
\textit{delta pose}
\( \displaystyle\Delta\mathbf{X}_{H_t}= \mathbf{X}_{H_0}^{-1}\mathbf{X}_{H_t}^{\phantom{*}}\) (relative to a fixed initial frame, e.g., average of initial poses from demonstrations) and  
\textit{relative pose}  
\( \displaystyle\delta\mathbf{X}_{H_t}= \mathbf{X}_{H_{t-1}}^{-1}\mathbf{X}_{H_t}^{\phantom{*}} \).  
Substituting these into Eq.~\eqref{eq:abs_pose_mirroring} yields
\begin{equation}
    \Delta\mathbf{X}_{H}^* = \mathbf{E}\,\Delta\mathbf{X}_{H}^{\phantom{*}}\,\mathbf{E},
    \quad
    \delta\mathbf{X}_{H}^* = \mathbf{E}\,\delta\mathbf{X}_{H}^{\phantom{*}}\,\mathbf{E},
    \label{eq:mirror_relative}
\end{equation}
which removes the dependency on the camera frame.
If an \textit{eye-in-hand} camera is rigidly attached to the end-effector, Eq.~\eqref{eq:abs_pose_mirroring} no longer applies in the global frame. 
However, the same symmetry holds in local coordinates, and the absolute mirrored pose can then be recovered by composition (see App.~A for derivation).
The associated actions are
\( \Delta\mathbf{a}_t := \Delta\mathbf{X}_{H_{t+1}} \) and
\( \delta\mathbf{a}_t := \delta\mathbf{X}_{H_{t+1}} \).
This canonicalization re-centers every trajectory starting at the identity and is thus dataset-agnostic.

\textbf{Discontinuity of Mirroring in \(\mathrm{SO(3)}\).}
To ensure the robot acquires the mirrored skill starting from a near-identical configuration, it is desirable for the initial end-effector pose to remain consistent before and after mirroring. 
However, in general, the mirrored initial pose satisfies \( \mathbf{X}_{H_0}^* \neq \mathbf{X}_{H_0} \), due to discontinuities introduced by reflection in \(\mathrm{SO(3)}\).
For brevity, we omit the local-frame markers (\(\delta,\,\Delta\)) in Eq.~\eqref{eq:mirror_relative} and overload
\(\mathbf{E}\!=\!\mathrm{diag}(-1,1,1)\) to act on \(3{\times}3\) matrices.  
Represent a global pose \(\mathbf{X}\in\mathrm{SE}(3)\) by rotation \(\mathbf{R}\in\mathrm{SO}(3)\) and translation \(\mathbf{t}\in\mathbb{R}^3\).
Reflection about the \(yz\)-plane maps
$
(\mathbf{R},\mathbf{t})
\;\longmapsto\;
(\mathbf{R}^*,\mathbf{t}^*)$
with
$
\mathbf{R}^*=\mathbf{E}\mathbf{R}\mathbf{E},\;
\mathbf{t}^*=\mathbf{E}\mathbf{t}.
$
Translations are altered mildly after the local reparameterization around the origin (\(\mathbf{t}^*\!\approx\!\mathbf{t}\) for small motions), whereas rotations generally are not:  
\(\mathbf{R}^*\) flips the first column and first row of \(\mathbf{R}\), producing an abrupt change (see App.~A for details). The phenomenon is avoided only when the end-effector’s \(x\)-axis already aligns with the world \(X\)-axis (\(\mathbf{R}_{(:,1)}\!\approx\![1,0,0]^\top\)), in which case \(\mathbf{R}^*\!\approx\!\mathbf{R}\).  
MirrorDuo enforces this condition via a constant alignment rotation 
\(\mathbf{Q}\), derived once from the average initial rotations across the dataset, 
which maps the mean tool axis \(\overline{\mathbf{R}}_{H_0}(:,1)\) to the world \(X\)-axis 
\(\mathbf{e}_x=[1,0,0]^\top\).

\textbf{The Dual Realization.}
The above formulation offers flexibility by enabling either data augmentation to enrich the distribution or geometric constraints that embed the inductive bias directly into the network architecture.
We first cast the geometric mirroring rule into a vector form that interfaces cleanly with neural networks.  
A pose \( \mathbf{X}=[\mathbf{R},\mathbf{t}]\in\mathrm{SE}(3) \) is vectorized as
$
\operatorname{vec}_X(\mathbf{X})
\;=\;
\bigl[\;\mathbf{t}^\top\; \mathbf{r}_1^\top\; \mathbf{r}_2^\top\bigr]^\top\!\in\mathbb{R}^9,
$
\(\mathbf{r}_{1,2}\) are the first two rotation columns~\cite{zhou2018continuity}.  
Eq.~\eqref{eq:abs_pose_mirroring} simplifies to:
\begin{equation}
    \operatorname{vec}_X\!\bigl({}^C\!\mathbf{X}_{H_t}^*\bigr)
    =\;
    \boldsymbol{\rho}\,\odot\,
    \operatorname{vec}_X\!\bigl({}^C\!\mathbf{X}_{H_t}\bigr),
    \quad
    \boldsymbol{\rho}\;=\;[-1,\;1^3,\;-1^3,\;1^2]^\top,
    \label{eq:mirror_vec_form}
\end{equation}
where \(1^n\) repeats the scalar 1 \(n\) times and \(\odot\) is element-wise multiplication. 
From Eq.~\eqref{eq:mirror_vec_form}, we define a unified mirroring operator \( \mathcal{M} \) over observation–action pairs:
\begin{equation}
  \mathcal{M}(\mathbf{o}_t) := \left(\mathcal{M}_I(\{\mathbf{I}_t^{(c)}\}),\ \boldsymbol{\rho} \odot \mathrm{vec}_X(\mathbf{s}_t)\right), \quad
\mathcal{M}(\mathbf{a}_t) := \boldsymbol{\rho} \odot \mathrm{vec}_X(\mathbf{a}_t),  
\label{eq:augmentation_notation}
\end{equation}
where \( \mathcal{M}_I(\cdot) \) denotes horizontal flipping in image space.

The \textit{data augmentation} variant of \textsc{MirrorDuo} operates during batch sampling without modifying the model. Given a minibatch of \( N \) trajectories, we randomly sample \( m \) trajectories and mirror their images, proprioception, and actions according to Eq.~\eqref{eq:augmentation_notation}. Mirrored samples replace the originals, yielding a mixture of original and mirrored data in each batch. Throughout all experiments in this paper, we set \( m/N \approx 0.5 \).
For brevity, we refer to MirrorDuo Data Augmentation as \textsc{MirrorAug}.

\textit{Mirror-equivariant diffusion policy\label{sec:mirror_diffusion}} embeds the mirror transformations as an inductive bias within the latent diffusion process, promoting consistency between original and mirrored samples. 
This approach leverages the strengths of diffusion models while preserving the introduced reflection-equivariance properties introduced by MirrorDuo.
We refer to it as \textsc{MirrorDiffusion}.

Diffusion models for behavior cloning~\cite{chi2023diffusionpolicy} train the noise prediction function \( \varepsilon_{\theta}(\mathbf{o}, \mathbf{a}^k + \varepsilon^k, k) \) to infer the noise \( \varepsilon^k \) added to the action at each forward diffusion step \( k \), based on the observation \( \mathbf{o} \) and noisy action prediction \( \mathbf{a}^k + \varepsilon^k \), where \( \mathbf{\theta }\) denotes the learnable weights and the noise schedule governs the injection of \( \varepsilon^k \).
During inference, starting from a noisy action \( \mathbf{a}^k \sim \mathcal{N}(\mathbf{0}, \mathbf{I}) \), the denoised action \( \mathbf{a}^0 \) is obtained iteratively via
\begin{equation}
   \mathbf{a}^{k-1} = \alpha \left( \mathbf{a}^k - \gamma \, \varepsilon_\theta(\mathbf{o}, \mathbf{a}^k, k) + \eta^k \right),
\quad \text{where} \quad \eta^k \sim \mathcal{N}(0, \sigma^2 \mathbf{I}), 
\label{eq:denoise}
\end{equation}
with \( \alpha \), \( \gamma \), and \( \sigma \) being functions of \( k \), governed by the noise schedule.
The equivariance of the diffusion process lies in its noise prediction function. A function \( f \) is equivariant to a group \( G \) if it commutes with the group transformations, i.e.,
$f\left( \rho_x(g) x \right) = \rho_y(g) f(x), \quad \forall g \in G$, where \( \rho_x \) and \( \rho_y \) are group representations mapping each group element \( g \) to an \( n \times n \) invertible transformation matrix acting on the input and output spaces ($x,\ y$), respectively. For brevity, we omit explicit notation for group representations.
As shown in~\cite{yang_equibot_sim3, wang_equivariant_diff_po}, the  per-step ground-truth noise prediction function \( \varepsilon\) is equivariant if the underlying policy \( \pi: \mathbf{o} \mapsto \mathbf{a} \) is equivariant:
$
\varepsilon(g \mathbf{o}, g \mathbf{a}^k, k) = g \, \varepsilon(\mathbf{o}, \mathbf{a}^k, k).
$

In this work, we enforce the reflection symmetry of the policy \( \pi \) by using \( \mathrm{E}(n) \)-equivariant Steerable CNNs~\cite{cesa2022_En_CNN}. The sign pattern in the representation matrix \( \boldsymbol{\rho} \) in Eq.~\eqref{eq:mirror_vec_form} corresponds to a block decomposition aligned with the irreducible representations of the reflection group acting on both the action and proprioceptive state spaces. Specifically,
$
\pi_{\boldsymbol{\xi}}(\mathcal{M}(\mathbf{o})) = \mathcal{M}(\pi_{\boldsymbol{\xi}}(\mathbf{o})),
$
where \( \boldsymbol{\xi} \) denotes the learnable parameters of the policy. Detailed network structure is in App.~B.
Note that the denoising function is per-step equivariant. However, the independent noise term \( \eta^k \) injected at each iteration in Eq.~\eqref{eq:denoise} introduces diversity, breaking global reflection symmetry in the reverse diffusion process. Removing this noise collapses the process to a single deterministic output~\cite{ho2020denoising}. We quantify the impact of this symmetry violation on task performance in Sec.~\ref{sec:visual_asym_exps}.

\textbf{Generalization under Visual Asymmetry.}
The goal of MirrorDuo is to synthesize immediately deployable mirrored trajectories without altering the robot’s initial state or relying on camera extrinsics. 
In the state space, this is achieved through local parameterization and centering at the fixed point of the reflection operator. 
In the image space, the analogous operation centers the end-effector on the vertical flip axis when the third-person camera is off-centered, either by estimating the camera pose via hand–eye calibration or by using vision models such as Grounded-SAM~\cite{ren2024grounded} for direct localization. 
Simulation experiments are in App.~F.

Furthermore, the above reflection symmetry formulations do not account for common sources of visual asymmetry under image mirroring, such as non-uniform table textures, background patterns, or asymmetries in the robot’s design. 
As shown in Fig.~\ref{fig:visual_asymm_from_robot}, the robot’s wrist appears left-sided in the mirrored view despite being consistently right-sided in the real world. 
Such discrepancies introduce visual out-of-distribution (OOD) artifacts.
To address this, we complement MirrorDuo with simple yet effective generalization techniques that mitigate mild violations of image reflection symmetry. 
As discussed in Sec.~\ref{sec:visual_asym_exps}, MirrorDuo’s tolerance to visual asymmetry is closely tied to the policy’s visual encoder and its capacity to generalize to task-irrelevant variations, such as lighting, distractors, and background differences.
We use a ResNet-18~\cite{he2016deep} pretrained on ImageNet~\cite{imagenet_cvpr09}, which has shown strong OOD robustness in manipulation~\cite{burns_resnet_ood}, and apply Random Overlay~\cite{hansen2021_soda,zhuang2024robosaga} to further enhance the visual robustness of the policies.

\section{Experiments}

\subsection{Simulation}
We design three evaluation settings to study different aspects of MirrorDuo. 
All tasks are based on the publicly available MimicGen dataset~\cite{mandlekar2023mimicgen}, with detailed task descriptions provided in App.~C. 
All experiments include both an eye-in-hand and a third-person camera. 
For clarity, we use the term \emph{mirrored demonstrations} to denote synthetic data generated through mirroring, and \emph{opposite-side demonstrations} to denote real data collected directly from the mirrored workspace.

\begin{figure}[t]
  \centering
  \subfloat[Close-view, one-side\label{fig:sim_close_one}]{
    \includegraphics[height=0.155\linewidth]{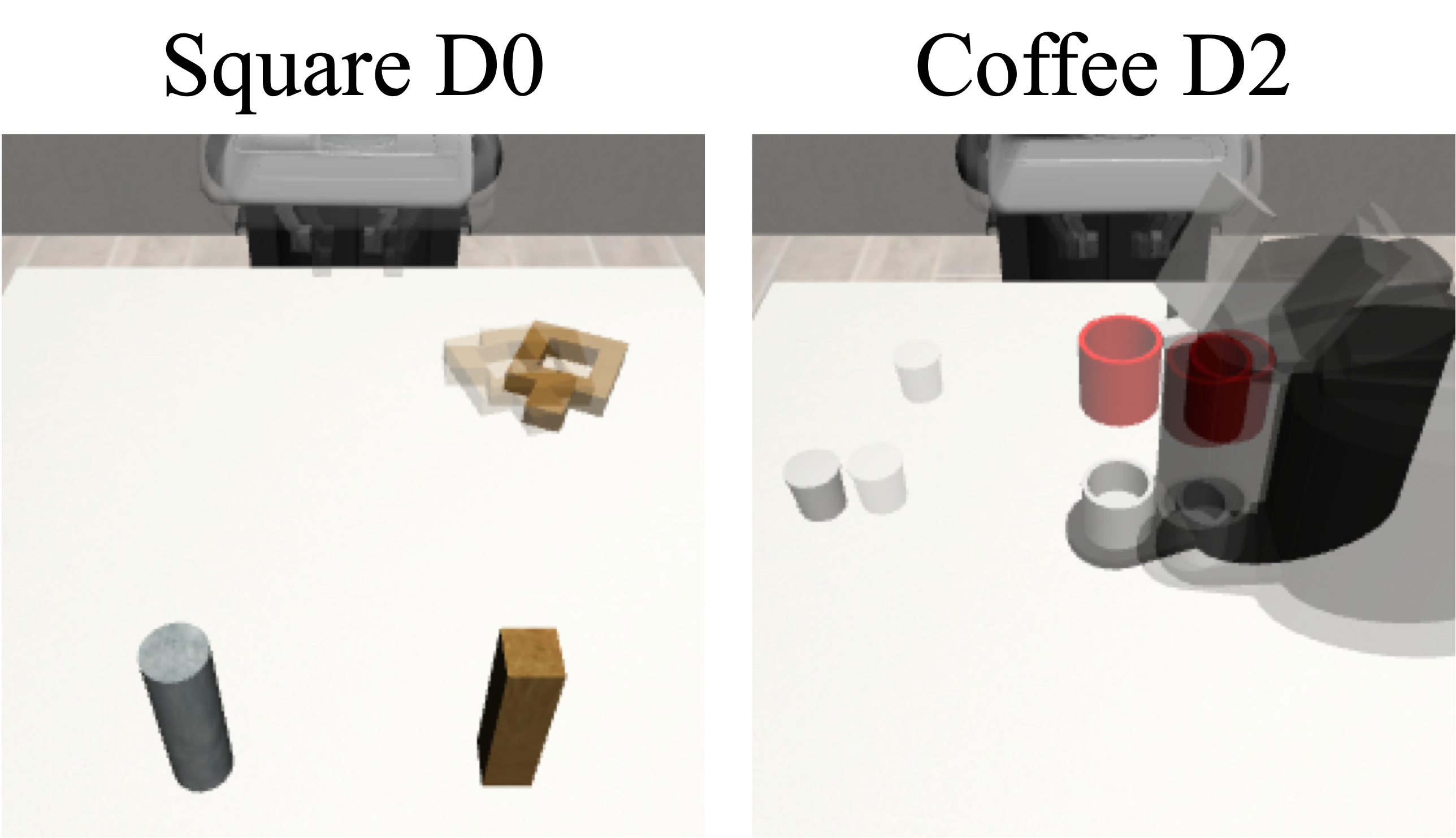}
  }
  \hfill
  \subfloat[Wide-view, one-side\label{fig:sim_wide_one}]{
    \includegraphics[height=0.155\linewidth]{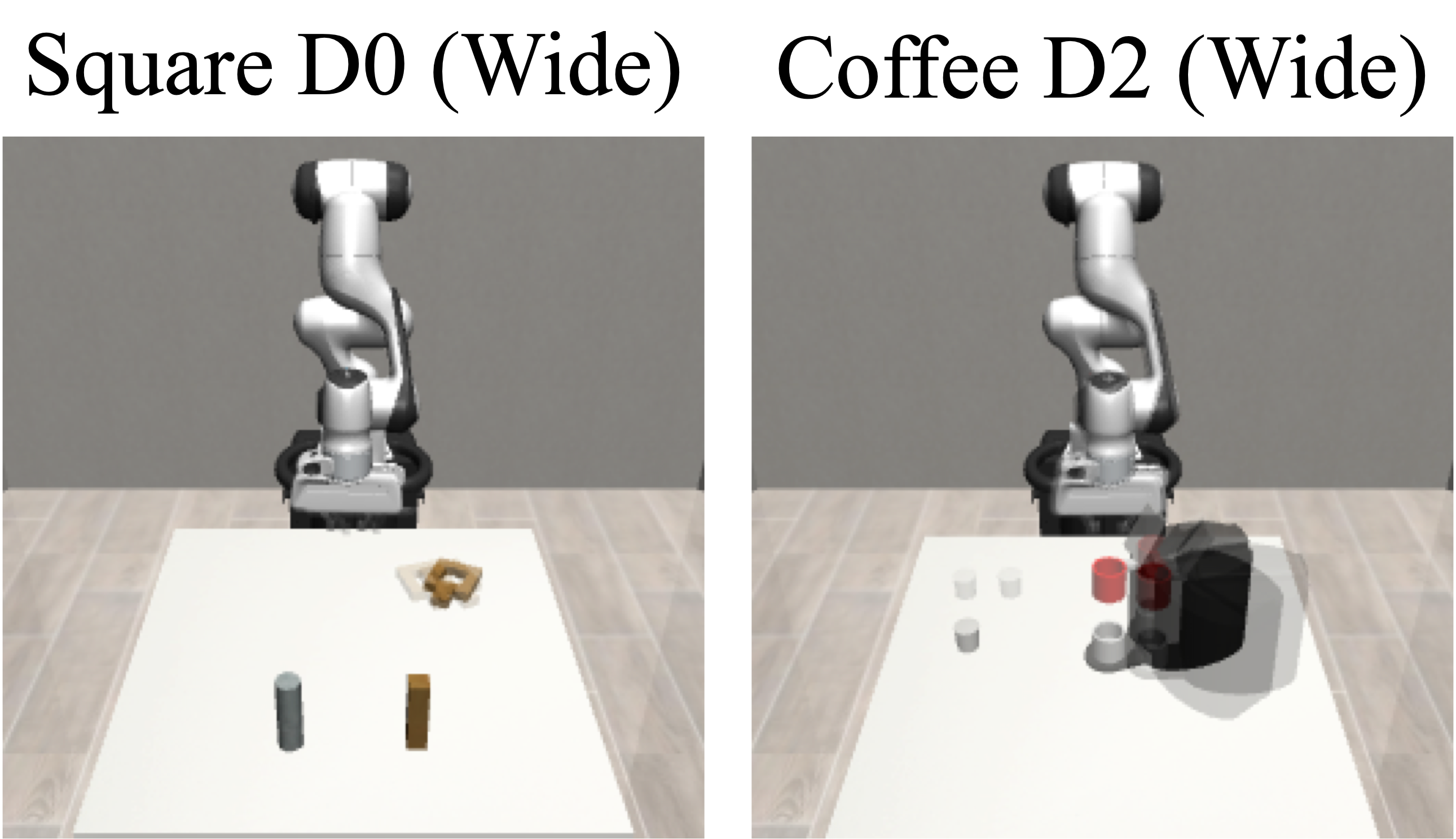}
  }
  \hfill
  \subfloat[Varying demonstrations, two-sides\label{fig:sim_two_sides}]{
    \includegraphics[height=0.155\linewidth]{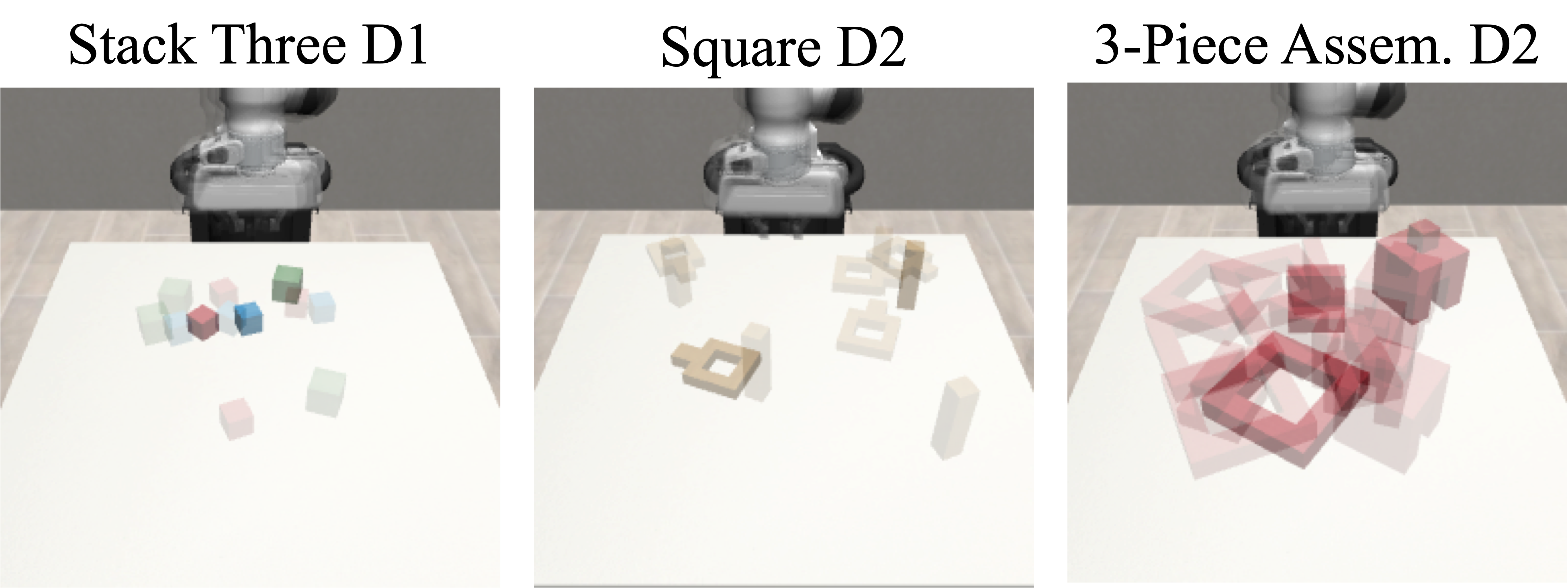}
  }

\caption{\textbf{Simulation Setups.} Each image shows the environment, averaged over several initial conditions.  
(a) Close-view with demonstrations confined to one half of the workspace.  
(b) Wide-view with the camera moved back.
(c) Intermediate-view with demos distributed across the tabletop.}
\label{fig:sim_setups}
\vspace{-2mm}
\end{figure}

\textbf{(I) Nearly Symmetric Visuals, One-side} (Fig.~\ref{fig:sim_close_one}).
For one-sided demonstrations, mirrored images differ slightly around the wrist and gripper compared to the true appearance from the opposite side (Fig.~\ref{fig:mirrored_close_view_ood}).
We evaluate the direct transferability of policies from the demonstration domain to the unseen mirrored setup, and assess improvements from integrating visual generalization techniques.

\textbf{(II) Visual Asymmetry from the Robot and Backgrounds, One-side} (Fig.~\ref {fig:sim_wide_one}). 

\begin{figure}[t]
  \centering
\includegraphics[width=0.98\linewidth]{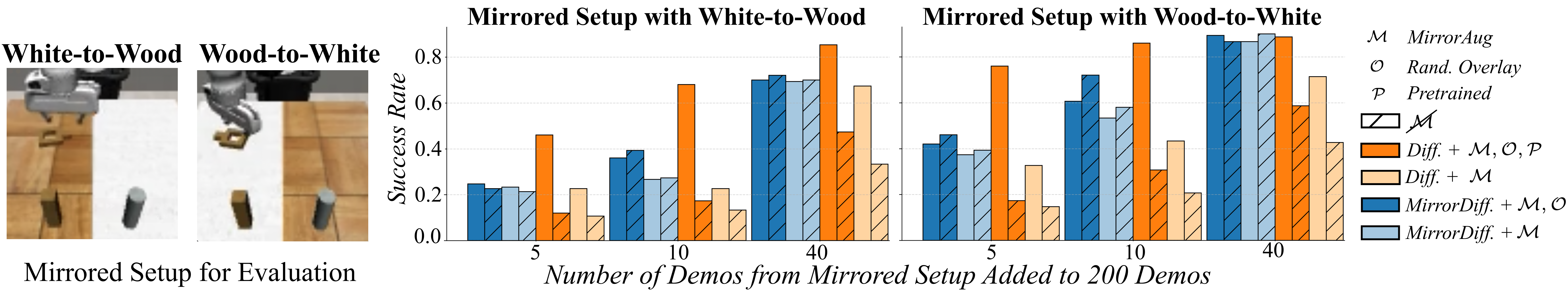}
\caption{\textbf{Setup and success rate (\%) for asymmetric backgrounds}, in the mirrored arrangements, against the number of additional opposite-side demos. Evaluation of \textit{Square D0} under mirrored setups with local visual domain shifts: white-to-wood and wood-to-white table textures.
}
\vspace{-5mm}
\label{fig:asymm_results}
\end{figure}

Moving the camera farther back reveals the robot's elbow, shoulder, and wrist, introducing stronger visual asymmetry compared with close-view (Fig.~\ref{fig:mirrored_wide_view_ood}). 
When table textures differ between sides, mirrored setups introduce background asymmetry (Fig.~\ref{fig:asymm_results}).
These settings are especially challenging given the higher visual complexity and limited data.
We study whether the reflection formulation combined with visual generalization can improve transferability under non-trivial asymmetric visual shifts, and how performance gains with varying numbers of opposite-side demos added. 

\textbf{(III) Two-sided Demonstrations for Long-horizon Tasks with Varying Data} (Fig.~\ref{fig:sim_two_sides}).
    Given the longer horizons and increased task complexity, with demonstrations exhibiting left–right symmetry, we vary the number of demonstrations to quantify how mirroring improves data efficiency when the mirrored domain geometrically overlaps with the demonstrated domain.

\subsubsection*{Setting I: Nearly Symmetric Visuals, One-side}
\label{sec:close_view_one_side_exps}

\begin{figure}[t]
    \centering
    \begin{minipage}[b]{0.58\linewidth}
        \centering
        \centering
\scriptsize
\begin{tabular}{lcccc}
\toprule
 & \multicolumn{2}{c}{\textbf{Square D0}} & \multicolumn{2}{c}{\textbf{Coffee D2}} \\
\cmidrule(lr){2-3} \cmidrule(lr){4-5}
& Original & Mirror & Original & Mirror \\
\midrule
& \multicolumn{4}{c}{\textit{MirrorDiffusion (Delta)}} \\
\cmidrule(lr){2-5}
$\cancel{{\mathcal{M}}}$ & 92$_{\pm 1}$ & 0$_{\pm 0}$ & 69$_{\pm 1}$ & 0$_{\pm 0}$ \\
$\mathcal{M}$ & 90$_{\pm 0}$ & 54$_{\pm 3}$ & 67$_{\pm 0}$ & 26$_{\pm 4}$ \\
$\mathcal{M},\mathcal{O}$ & \textbf{92}$_{\pm 0}$ & \textbf{64}$_{\pm 1}$ & \textbf{72}$_{\pm 2}$ & \textbf{28}$_{\pm 1}$ \\
\cmidrule(lr){1-5}
& \multicolumn{4}{c}{\textit{Diffusion Policy (Delta)}} \\
\cmidrule(lr){2-5}
$\mathcal{M}$ & 86$_{\pm 0}$ & 46$_{\pm 0}$ & \textbf{66}$_{\pm 3}$ & 21$_{\pm 0}$ \\
$\mathcal{M},\mathcal{O}$ & 83$_{\pm 1}$ & 79$_{\pm 2}$ & \textbf{66}$_{\pm 6}$ & \textbf{35}$_{\pm 2}$ \\
$\mathcal{M},\mathcal{O},\mathcal{P}$ & \textbf{95}$_{\pm 0}$ & \textbf{93}$_{\pm 0}$ & 62$_{\pm 0}$ & 33$_{\pm 3}$ \\
\cmidrule(lr){1-5}
& \multicolumn{4}{c}{\textit{BC-RNN (Relative)}} \\
\cmidrule(lr){2-5}
$\mathcal{M}$ & 71$_{\pm 1}$ & 3$_{\pm 0}$ & 50$_{\pm 1}$ & 0$_{\pm 0}$ \\
$\mathcal{M},\mathcal{O}$ & 80$_{\pm 2}$ & 42$_{\pm 3}$ & 54$_{\pm 2}$ & \textbf{32}$_{\pm 3}$ \\
$\mathcal{M},\mathcal{O},\mathcal{P}$ & \textbf{82}$_{\pm 0}$ & \textbf{53}$_{\pm 4}$ & \textbf{61}$_{\pm 4}$ & 17$_{\pm 2}$ \\
\bottomrule
\end{tabular}

    \captionof{table}{
        \textbf{Success rate (\%) for close-view, one-side demos.}
        Policies are trained on 200 \emph{Original} with their preferred action representations, and evaluated on both Original and Mirrored setups.
        $\cancel{\mathcal{M}}$ / $\mathcal{M}$: MirrorAug disabled / enabled,
        $\mathcal{O}$: Random Overlay,
        $\mathcal{P}$: Pretrained.
    }
    \label{tab:settings_1}
    \end{minipage}
    \hfill
    \begin{minipage}[b]{0.38\linewidth}
        \centering
        \subfloat[Close-view\label{fig:mirrored_close_view_ood}]{
            \includegraphics[width=0.95\linewidth]{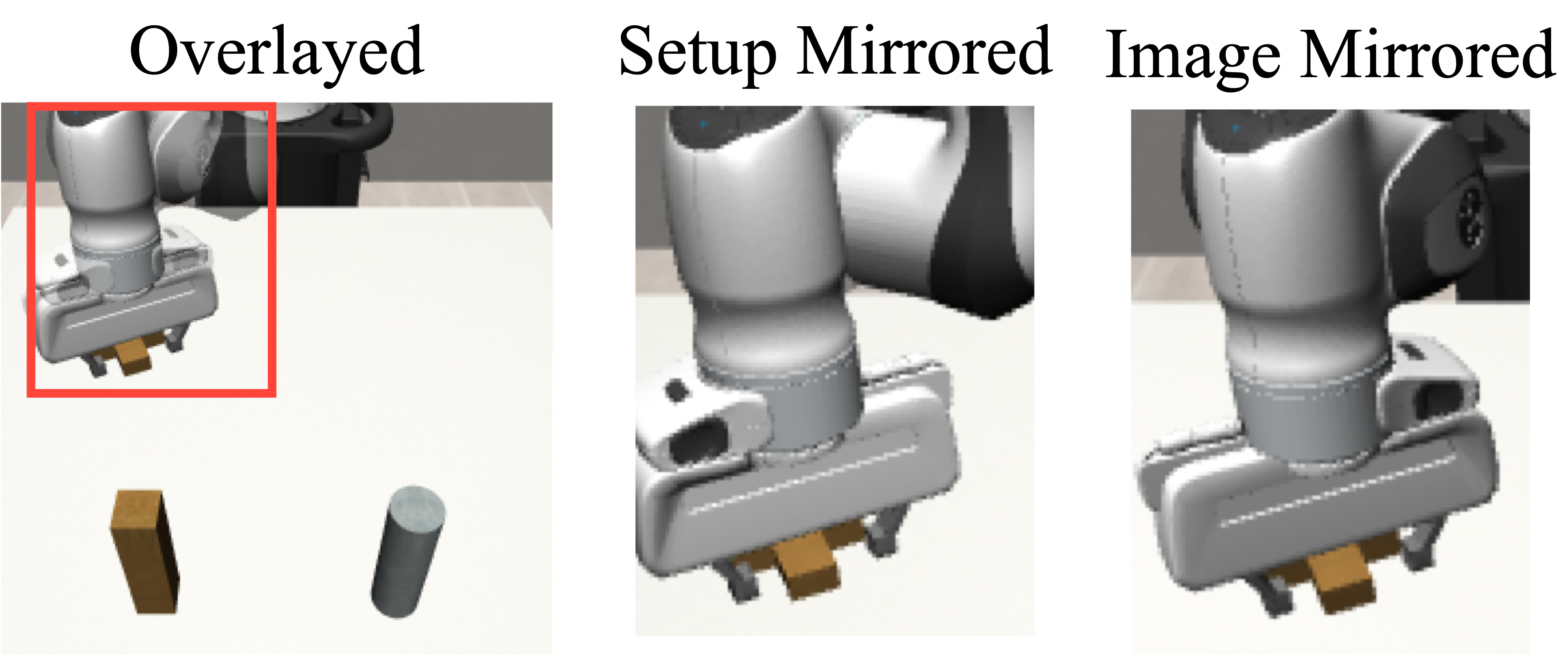}
        }
        \\
        \subfloat[Wide-view\label{fig:mirrored_wide_view_ood}]{
            \includegraphics[width=0.95\linewidth]{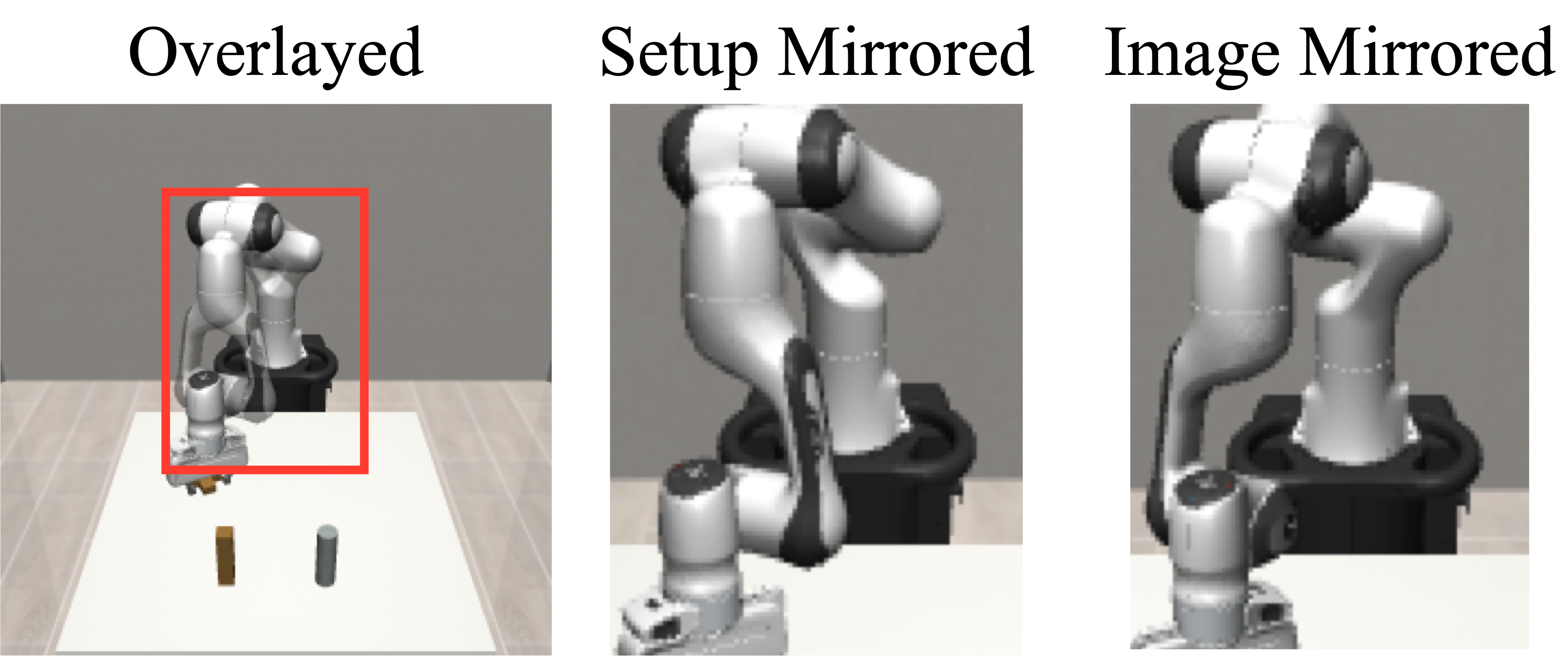}
        }
        \captionof{figure}{
        \textbf{Visual asymmetry from the robot.} 
        In the close view, asymmetry appears near the wrist and gripper, while in the wide view it extends to the elbow and shoulder.
        }
        \label{fig:visual_asymm_from_robot}
    \end{minipage}
    \vspace{-5mm}
\end{figure}

We evaluate three policy variants: MirrorDiffusion as described in Sec.~\ref{sec:mirror_diffusion}, Diffusion Policy~\cite{chi2023diffusionpolicy}, and BC-RNN~\cite{robomimic2021}, each combined with MirrorAug ($\mathcal{M}$), Random Overlay~\cite{zhuang2024robosaga} ($\mathcal{O}$), and, when available, pretrained visual backbone weights ($\mathcal{P}$).
Diffusion Policy and BC-RNN use ResNet-18 backbones initialized from ImageNet-pretrained weights, fine-tuned with a backbone learning rate one-tenth that of the policy head. MirrorDiffusion uses randomly initialized backbones, as no off-the-shelf pretrained weights exist for reflection-equivariant architectures. 
Evaluation is performed on original and mirrored object arrangements, with success rates averaged over the top three rollouts, sampled every 10 epochs across 250 training epochs.

Tab.~\ref{tab:settings_1} shows that \textbf{direct} transferability varies across policies and tasks. On \textit{Square D0}, Diffusion Policy with Random Overlay and  pretrained visual backbones matches its in-domain performance at 93\%. 
In contrast, on \textit{Coffee D2}, all methods perform similarly with transfer success rates around 30\%. MirrorDiffusion without MirrorAug fails entirely, achieving 0\% direct transfer. As detailed in Sec.~\ref{sec:mirror_diffusion}, although its denoising function is per-step equivariant, reflection symmetry is broken over the whole denoising trajectory due to independently sampled noise at each reverse step (Eq.~\eqref{eq:denoise}).

While no clear winner emerges between using Random Overlay alone versus combining it with pretrained weights, improved visual robustness consistently correlates with better direct transferability across tasks and policies. 
BC-RNN shows marked gains with mirrored success rates rise from near zero with MirrorAug alone to 53\% on \textit{Square D0} and 32\% on \textit{Coffee D2}.

\begin{figure}[t]
  \centering
\includegraphics[width=0.9\linewidth]{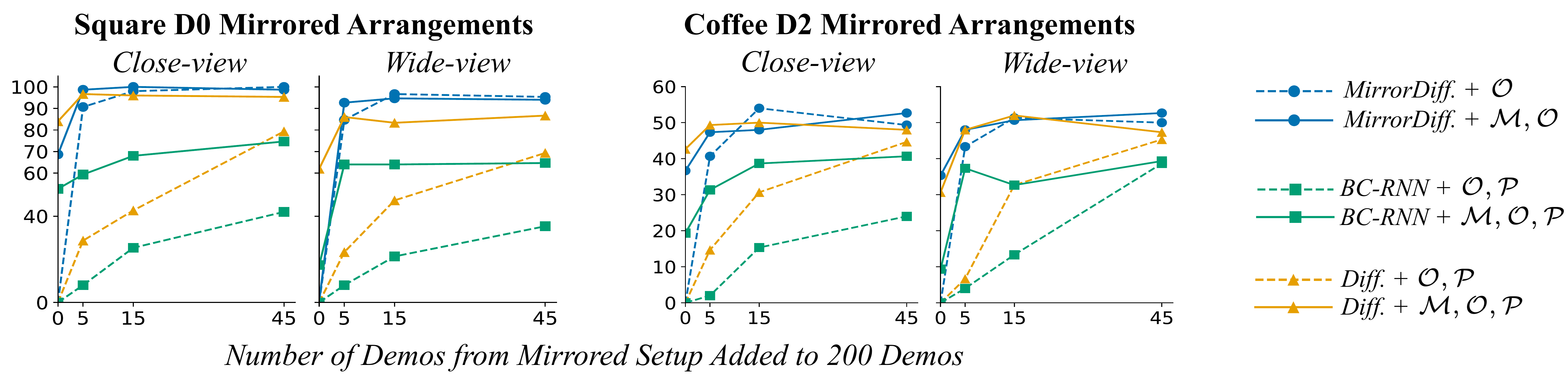}
\caption{\textbf{Wide-view success rate (\%)}, against number of additional opposite-side demos.
}
\vspace{-3mm}
\label{fig:data_efficiency}
\end{figure}

\begin{table}[t]
    \centering
\scriptsize
\definecolor{gainblue}{RGB}{58,147,195}
\definecolor{gainred}{RGB}{215,95,76}
\setlength{\tabcolsep}{4pt}
\begin{tabular}{l@{\hskip 4pt}lccc ccc ccc}
\toprule
& & \multicolumn{3}{c}{\textbf{Stack Three D1}} & \multicolumn{3}{c}{\textbf{Square D2}} & \multicolumn{3}{c}{\textbf{3-Part Assembly D2}} \\
\cmidrule(lr){3-5} \cmidrule(lr){6-8} \cmidrule(lr){9-11}
& \textbf{Method} & \textit{50} & \textit{100} & \textit{200} & \textit{100} & \textit{200} & \textit{500} & \textit{100} & \textit{200} & \textit{500} \\
\midrule
\multirow{6}{*}{\rotatebox[origin=c]{90}{\textbf{Delta}}}
& EquiDiff. $+\, \mathcal{O}$ & 20.7 & 54.7 & 77.3 & 25.3 & 41.3 & 60.0 & 15.3 & 39.3 & 63.0\\
& MirrorDiff. $+\, \mathcal{O}$ & $51.3_{\textcolor{gainblue}{\uparrow31}}$ & $77.3_{\textcolor{gainblue}{\uparrow23}}$ & $89.3_{\textcolor{gainblue}{\uparrow12}}$ & $24.0_{\textcolor{gainred}{\downarrow1}}$ & $48.7_{\textcolor{gainblue}{\uparrow7}}$ & $59.3_{\textcolor{gainred}{\downarrow1}}$ & $25.3_{\textcolor{gainblue}{\uparrow10}}$ & $49.3_{\textcolor{gainblue}{\uparrow10}}$ & $61.3_{\textcolor{gainred}{\downarrow2}}$\\
\cmidrule(){2-11}
& DiffPo. $+\,\mathcal{O}\,\mathcal{P}$& 19.3 & 47.3 & 80.7 & 20.7 & 40.0 & 58.0 & 11.0 & 31.3 & 64.0\\
& DiffPo. $+\, \mathcal{M},\,\mathcal{O},\,\mathcal{P}$ & $50.7_{\textcolor{gainblue}{\uparrow31}}$ & $68.0_{\textcolor{gainblue}{\uparrow21}}$ & $91.3_{\textcolor{gainblue}{\uparrow11}}$ & $32.7_{\textcolor{gainblue}{\uparrow12}}$ & $49.3_{\textcolor{gainblue}{\uparrow9}}$ & $56.7_{\textcolor{gainred}{\downarrow1}}$ & $21.0_{\textcolor{gainblue}{\uparrow10}}$ & $47.3_{\textcolor{gainblue}{\uparrow16}}$ & $61.3_{\textcolor{gainred}{\downarrow3}}$\\
\cmidrule(){2-11}
& BC-RNN $+\, \mathcal{O},\,\mathcal{P}$ & 0.7 & 2.0 & 8.0 & 4.0 & 9.3 & 32.0 & 0.0 & 0.0 & 6.0\\
& BC-RNN $+\, \mathcal{M},\,\mathcal{O},\,\mathcal{P}$ & $0.0_{\textcolor{gainred}{\downarrow1}}$ & $3.3_{\textcolor{gainblue}{\uparrow1}}$ & $37.3_{\textcolor{gainblue}{\uparrow29}}$ & $4.7_{\textcolor{gainblue}{\uparrow1}}$ & $12.7_{\textcolor{gainblue}{\uparrow3}}$ & $43.3_{\textcolor{gainblue}{\uparrow11}}$ & $1.0_{\textcolor{gainblue}{\uparrow1}}$ & $3.3_{\textcolor{gainblue}{\uparrow3}}$ & $12.0_{\textcolor{gainblue}{\uparrow6}}$\\
\midrule
\multirow{6}{*}{\rotatebox[origin=c]{90}{\textbf{Relative}}}
& EquiDiff. $+\, \mathcal{O}$ & 6.7 & 25.3 & 62.7 & 11.3 & 20.7 & 40.0 & 1.3 & 4.7 & 22.0\\
& MirrorDiff. $+\, \mathcal{O}$& $28.7_{\textcolor{gainblue}{\uparrow22}}$ & $57.3_{\textcolor{gainblue}{\uparrow32}}$ & $80.0_{\textcolor{gainblue}{\uparrow17}}$ & $18.0_{\textcolor{gainblue}{\uparrow7}}$ & $32.0_{\textcolor{gainblue}{\uparrow11}}$ & $47.3_{\textcolor{gainblue}{\uparrow7}}$ & $13.3_{\textcolor{gainblue}{\uparrow12}}$ & $23.3_{\textcolor{gainblue}{\uparrow19}}$ & $50.0_{\textcolor{gainblue}{\uparrow28}}$\\
\cmidrule(){2-11}
& DiffPo. $+\, \mathcal{O},\,\mathcal{P}$ & 19.3 & 31.3 & 58.0 & 18.0 & 30.0 & 44.0 & 4.0 & 13.3 & 32.0\\
& DiffPo. $+\, \mathcal{M},\,\mathcal{O},\,\mathcal{P}$ & $29.3_{\textcolor{gainblue}{\uparrow10}}$ & $50.0_{\textcolor{gainblue}{\uparrow19}}$ & $68.0_{\textcolor{gainblue}{\uparrow10}}$ & $32.7_{\textcolor{gainblue}{\uparrow15}}$ & $41.3_{\textcolor{gainblue}{\uparrow11}}$ & $45.3_{\textcolor{gainblue}{\uparrow1}}$ & $11.0_{\textcolor{gainblue}{\uparrow7}}$ & $26.7_{\textcolor{gainblue}{\uparrow13}}$ & $48.0_{\textcolor{gainblue}{\uparrow16}}$\\
\cmidrule(){2-11}
& BC-RNN $+\, \mathcal{O},\,\mathcal{P}$ & 6.7 & 18.0 & 51.3 & 8.0 & 19.3 & 45.3 & 2.0 & 3.3 & 11.3\\
& BC-RNN $+\, \mathcal{M},\,\mathcal{O},\,\mathcal{P}$ & $18.0_{\textcolor{gainblue}{\uparrow11}}$ & $35.3_{\textcolor{gainblue}{\uparrow17}}$ & $73.3_{\textcolor{gainblue}{\uparrow22}}$ & $16.0_{\textcolor{gainblue}{\uparrow8}}$ & $24.7_{\textcolor{gainblue}{\uparrow5}}$ & $48.0_{\textcolor{gainblue}{\uparrow3}}$ & $1.0_{\textcolor{gainred}{\downarrow1}}$ & $8.7_{\textcolor{gainblue}{\uparrow5}}$ & $22.7_{\textcolor{gainblue}{\uparrow11}}$\\
\bottomrule
\end{tabular}
    \vspace{1mm}
\caption{\textbf{Setting III: Wide-view, two-side demonstrations.} Success rate (\%) on three MimicGen tasks as the number of demos increases. Blue denotes absolute gains from MirrorAug or MirrorDiffusion over their baselines; red indicates drops. \textit{Delta} and \textit{Relative} refer to action controllers. Results are averaged over the top-1 rollout from three training seeds. Full table in App.~D.}
    \label{tab:main_results}
    \vspace{-8mm}
\end{table}

\subsubsection*{Setting II: Visual Asymmetry from the Robot and Backgrounds, One-side}
\label{sec:visual_asym_exps}
To assess MirrorDuo under non-trivial visual asymmetry in one-sided tasks, we evaluate \textit{Square D0} and \textit{Coffee D2} with wide-view images, and \textit{Square D0} with asymmetric table textures. 
In both cases, opposite-side demonstrations are generated via Eq.~\eqref{eq:abs_pose_mirroring} and validated through successful executions. 
MirrorDiffusion, Diffusion Policy, and BC-RNN are tested with combinations of MirrorAug (\(\mathcal{M}\)), Random Overlay (\(\mathcal{O}\)), and pretrained weights (\(\mathcal{P}\); unavailable for MirrorDiffusion).

With \textbf{wide-view}, all demos are re-rendered with a wide camera view, exposing asymmetries such as the elbow and shoulder links, making zero-shot transfer considerably harder than in close-view (Fig.~\ref{fig:visual_asymm_from_robot}).  
As shown in Fig.~\ref{fig:data_efficiency}, for \textit{Square D0}, the direct transfer performance of Diffusion Policy (\(\mathcal{M,\,O,\,P}\)) drops by approximately 30\%, while BC-RNN (\(\mathcal{M,\,O,\,P}\)) and MirrorDiffusion (\(\mathcal{M,\, O}\)) fall below 20\%, with BC-RNN dropping to zero.
Fig.~\ref{fig:data_efficiency} also demonstrates that policies with MirrorAug (solid lines) recover near in-domain performance with as few as 5 opposite-side demos, gaining at most after adding 10 opposite demos and plateauing beyond 15.
In contrast, without MirrorAug (green and yellow dashed lines), Diffusion Policy and BC-RNN require substantially more data to achieve comparable performance. 
Notably, although MirrorDiffusion (\(\mathcal{O}\)) fails at direct transfer due to symmetry violations introduced by step-wise noise during reverse diffusion, it matches the performance of MirrorDiffusion \((\mathcal{M},\, \mathcal{O}\)) with as few as additional 5 opposite-side demos, showing that structural reflection-equivariance can be effectively unlocked for workspace generalization.

We alternated half of the \textbf{table textures} (white/wood) in \textit{Square D0} and tested under mirrored setups with the opposite texture, where white-to-wood is notably harder due to higher visual complexity and limited demonstrations (Fig.~\ref{fig:asymm_results}). 
Without visual generalization (faded bars), MirrorDiff. and Diffusion \((\mathcal{M})\) both improve data efficiency over vanilla Diffusion, with MirrorDiff. showing a slight advantage. 
With the selected visual generalization techniques, MirrorDiff. \((\mathcal{O})\) gains modestly, while Diffusion \((\mathcal{M},\,\mathcal{O},\,\mathcal{P})\) benefits most, consistent with the gap observed between MirrorDiff. \((\mathcal{O})\) and Diffusion \((\mathcal{M},\,\mathcal{O},\,\mathcal{P})\) in real experiments (Sec.~\ref{sec:real_exps}). 
Overall, MirrorDiff. is preferred without generalization techniques, but with lightweight visual generalization methods, Diffusion \((\mathcal{M},\,\mathcal{O},\,\mathcal{P})\) achieves superior performance under visually challenging, limited-data conditions while also being faster. 
MirrorDiff., by contrast, is more computationally demanding due to its more sophisticated equivariant structure. 
We therefore recommend Diffusion \((\mathcal{M},\,\mathcal{O},\,\mathcal{P})\) as the practical choice.

\subsubsection*{Setting III: Two‐Side Demonstrations with Challenging Long Horizon Tasks}
\label{sec:two_side_exps}
We evaluate MirrorDuo's data efficiency with challenging long-horizon demonstrations covering both sides of the workspace (Fig.~\ref{fig:sim_two_sides}). 
Experiments are conducted on \textit{Stack Three D1}, \textit{Square D2}, and \textit{Three-piece Assembly D1}.
To match task difficulty, the number of demonstrations is varied: [50, 100, 200] for \textit{Stack Three D1}, where performance saturates quickly, and [100, 200, 500] for the more challenging \textit{Square D2} and \textit{Three-piece Assembly D1}.

Beyond the previously evaluated policies, we also compare against the $\mathrm{SO}(2)$-Equivariant Diffusion Policy~\cite{wang_equivariant_diff_po} as introduced in Sec.~\ref{sec:related_work}.

However, their relative controller is defined using global transformations, \(\mathbf{A}_t = \mathbf{X}_{H_t}^{\phantom{*}} \mathbf{X}_{H_{t-1}}^{-1}\), whereas applying our relative formulation (Eq.~\eqref{eq:mirror_relative}) would induce $\mathrm{SO}(2)$ invariance, differing from their original design. Nonetheless, the comparison remains valid since both delta and absolute controllers operate in fixed frames, and both relative formulations involve frame-to-frame transformations.
Detailed performance benchmarking is provided in Tab.~\ref{tab:avg_abs_improvement}.

\begin{wraptable}{b}{0.45\textwidth}
\centering
\scriptsize
\setlength{\tabcolsep}{3pt}
\begin{tabular}{llccc}
\toprule
 & \textbf{Method} & \makecell{\textbf{Low}\\\textbf{Demos}} & \makecell{\textbf{Medium}\\\textbf{Demos}} & \makecell{\textbf{High}\\\textbf{Demos}} \\
\midrule
\multirow{3}{*}{\rotatebox[origin=c]{90}{\textbf{Delta}}}
 & MirrorDiff. $+\, \mathcal{O}$ & 13.1 & 13.3 & 3.2 \\
 & DiffPo $+\, \mathcal{M},\,\mathcal{O},\,\mathcal{P}$ & 17.8 & 15.3 & 2.2 \\
 & BC-RNN $+\, \mathcal{M},\,\mathcal{O},\,\mathcal{P}$ & 0.3 & 2.7 & 15.6 \\
\midrule
\multirow{3}{*}{\rotatebox[origin=c]{90}{\textbf{Relative}}}
 & MirrorDiff. $+\, \mathcal{O}$ & 13.6 & 20.7 & 17.5 \\
 & DiffPo. $+\, \mathcal{M},\,\mathcal{O},\,\mathcal{P}$ & 10.6 & 14.4 & 9.1 \\
 & BC-RNN $+\, \mathcal{M},\,\mathcal{O},\,\mathcal{P}$ & 6.1 & 9.3 & 12.0 \\
\bottomrule
\end{tabular}
\caption{Average absolute gains over three tasks relative to each method’s baseline.}
\vspace{-2mm}
\label{tab:avg_abs_improvement}
\end{wraptable}

The absolute performance improvements over each method’s baseline are averaged across tasks, grouped by demonstration level (low, medium, high), and controller mode. 
MirrorDuo yields consistent gains in low- and medium-data regimes, with improvements plateauing once sufficient demonstrations are available. Notably, it significantly enhances performance under the less favorable controller for each policy, specifically, the relative controller for Diffusion and the delta controller for BC-RNN. However, BC-RNN continues to struggle under the relative setting, showing near-zero performance with or without MirrorAug when the data is insufficient.
We observe comparable performance between Diffusion \((\mathcal{M},\,\mathcal{O},\,\mathcal{P})\) and the explicitly reflection-equivariant MirrorDiffusion \((\mathcal{O})\) under delta control. 
This indicates that there is no substantial advantage between the two approaches under trivial visual asymmetry.

\subsection{Real-World Experiments}
\label{sec:real_exps}
\begin{figure}[t]
        \centering
        \subfloat[Task Distributions\label{fig:real_dist}]{
            \includegraphics[height=0.2\linewidth]{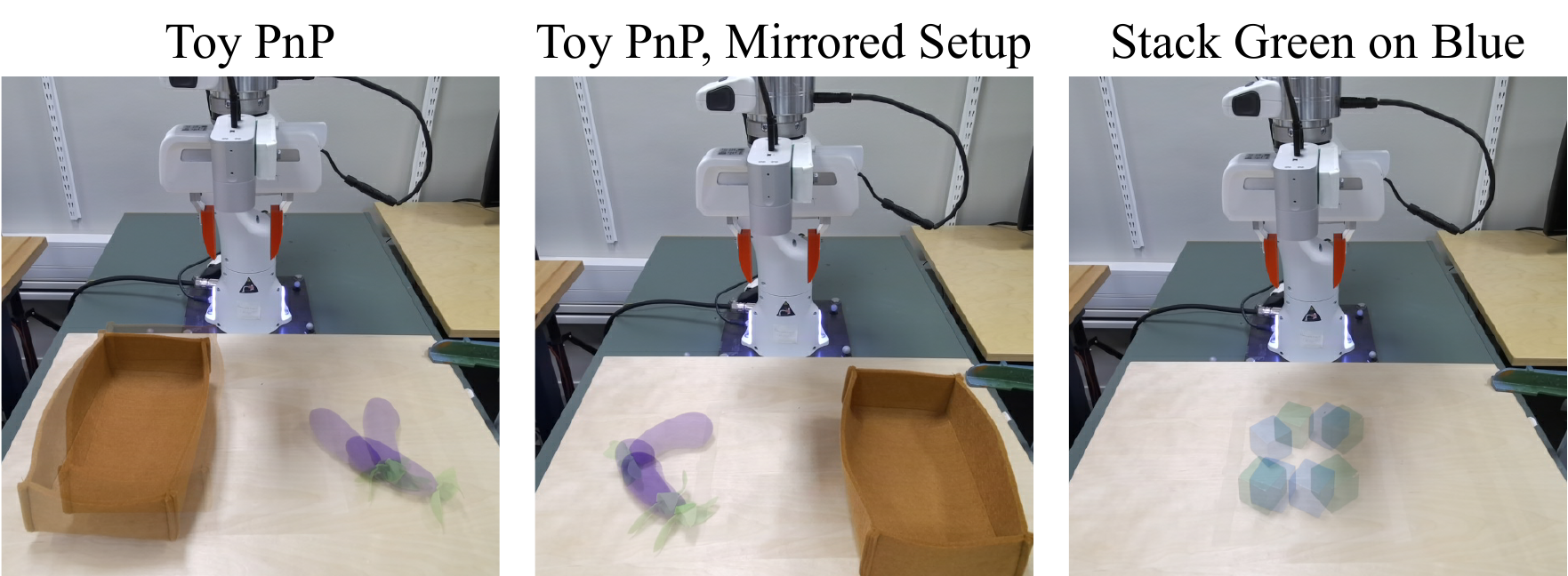}
        }
        \hspace{5mm}
        \subfloat[Task Goals\label{fig:real_goal}]{
            \includegraphics[height=0.2\linewidth]{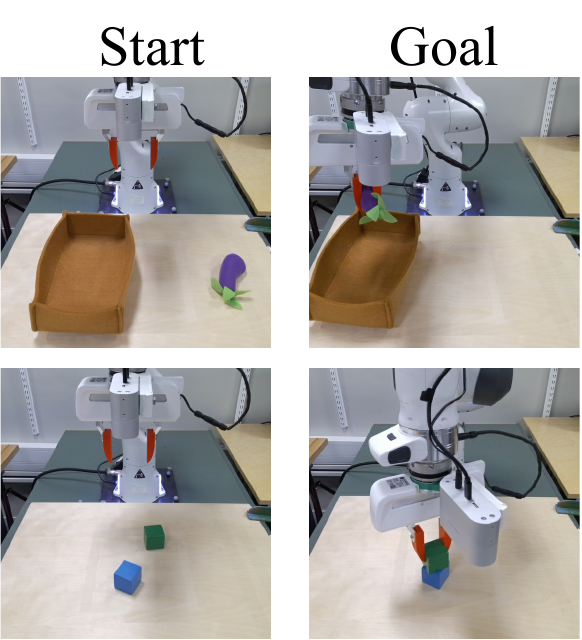}
        }
        \captionof{figure}{
        \textbf{Illstrations of Real Task Setups.}
        (a) Task Distribution: Each image overlay with three task arrangements. 
        (b) Example of start and goal configuration.
        }
        \label{fig:real_setup}
        \vspace{-2mm}
    \end{figure}
    
We evaluate on a real-world setup with a Franka Emika arm, a third-person Azure Kinect, and an eye-in-hand Orbbec Femto Bolt camera. 
Both cameras provide RGB-D input, though only RGB is used during training and inference. Demonstrations are collected via teleoperation~\cite{welle2024quest2ros} at 10~Hz.
We design two tasks as real-world counterparts of the one-side and two-side settings in simulation: (1) \textbf{One-side}: a pick-and-place task, where a plush toy is picked from the right and placed in a left-sided bin; and (2) \textbf{Two-side}: a block-stacking task, where a green block is stacked onto a blue one, with demonstrations evenly distributed across both sides. Objects are randomly rotated, and placement regions are shown in Fig.~\ref{fig:real_setup}. We evaluate the direct transferability and data efficiency of Diffusion Policy \((\mathcal{M},\,\mathcal{O},\,\mathcal{P})\) and MirrorDiffusion \((\mathcal{O})\). 
Training follows the same protocol as in simulation, with evaluation performed on the best checkpoint selected by validation loss in 400 epochs.

\begin{table*}[t]
    \centering
    \scriptsize
    \begin{minipage}[t]{0.48\linewidth}
        \centering
        \begin{tabular}{lccc}
            \toprule
            & \multirow{2}{*}{\textbf{In-domain}} & \multicolumn{2}{c}{\textbf{\# M-Demos}} \\
            \cmidrule(lr){3-4}
            & & 0 & 5 \\
            \midrule
            MirrorDiff. $+\, \mathcal{O}$ & 76.7 & 0.0 & 73.3 \\
            DiffPo. $+\, \mathcal{M},\,\mathcal{O},\,\mathcal{P}$ & \textbf{86.7} & \textbf{20.0} & \textbf{83.3} \\
            DiffPo. $+
        \,\mathcal{O},\,\mathcal{P}$ & 83.3 & 0.0 & 3.3 \\
            \bottomrule
        \end{tabular}
        \caption{\textbf{Generalization to Mirrored Setup.} Success rates (out-of 30 trials) on the plush toy task in original and mirrored setups. M-demos denote demonstrations in the mirrored setup.}
        \label{tab:real_transfer}
    \end{minipage}
    \hfill
    \begin{minipage}[t]{0.48\linewidth}
        \centering
        \begin{tabular}{lcc}
            \toprule
            & \multicolumn{2}{c}{\textbf{\# Demos}} \\
            \cmidrule(lr){2-3}
            & 200 & 300 \\
            \midrule
            MirrorDiff. $+\, \mathcal{
            O}$ & 43.0 &  60.0 \\
            DiffPo. $+\, \mathcal{M},\,\mathcal{O},\,\mathcal{P}$ & \textbf{66.7} & \textbf{73.3} \\
            DiffPo. $+\,\mathcal{O},\,\mathcal{P}$ & 40.0 & 53.3 \\
            \bottomrule
        \end{tabular}
        \caption{\textbf{In-domain Data Efficiency.} Success rates (averaged over 30 trials) for the block-stacking task, evaluated with increasing numbers of demonstrations.
        }
        \label{tab:data_efficiency}
    \end{minipage}
    \vspace{-5mm}
\end{table*}

As shown in Tab.~\ref{tab:real_transfer}, with only 5 opposite-side demonstrations, both MirrorDiffusion \((\mathcal{O})\) and Diffusion Policy \((\mathcal{M},\,\mathcal{O},\,\mathcal{P})\) effectively transfer to the mirrored setup, achieving 73.3\% and 83.3\%, respectively, matching their in-domain performance.
This is not observed for the standard Diffusion Policy, when trained primarily on one-side demonstrations, achieves only 3.3\% in the mirrored domain.
In the stacking task, Diffusion Policy \((\mathcal{M},\,\mathcal{O},\,\mathcal{P})\)  achieves gains of 26.7\% and 20\% over the baseline Diffusion Policy.
Similar to Setting II (Sec.~\ref{sec:visual_asym_exps}) in simulation, MirrorDiffusion \((\mathcal{O})\) lags behind Diffusion Policy \((\mathcal{M},\,\mathcal{O},\,\mathcal{P})\) under challenging visual asymmetry, performing only marginally better than the baseline Diffusion Policy.
In the plush toy task, Diffusion Policy \((\mathcal{M},\,\mathcal{O},\,\mathcal{P})\)  outperforms MirrorDiffusion \((\mathcal{O})\) by approximately 10\% in both the original and mirrored domains. 
This may be attributed to the use of pretrained visual backbones, which offer improved robustness to real-world variations such as slight shadows, along with faster convergence during training.

In summary, while MirrorDiffusion is favored without visual generalization techniques, Diffusion \((\mathcal{M},\,\mathcal{O},\,\mathcal{P})\) achieves higher performance under challenging, limited-data conditions and offers more efficient training. 
We therefore recommend Diffusion Policy \((\mathcal{M},\,\mathcal{O},\,\mathcal{P})\) as the practical choice.

\section{Conclusion}
\label{sec:conclusion}
We introduced MirrorDuo, a general framework for leveraging reflectional symmetry in image-based visuomotor policy learning. 
By mirroring RGB observations, 6-DoF proprioception, and actions in a semantically consistent way, our approach enables two complementary applications: MirrorAug, a data augmentation strategy that expands training coverage to mirrored configurations, and MirrorDiffusion, a reflection-equivariant policy that generalizes by construction. 
Despite the complexity of projecting 3D reflections into 2D visual observations, our results demonstrate that reflectional symmetry is highly compatible with image-based learning pipelines. 
Combined with lightweight visual generalization techniques, policies trained with MirrorDuo generalize effectively to mirrored setups with minimal or no additional opposite-side data and are tolerant to visual asymmetries introduced by both the robot and the background (e.g., table textures). 
Together, these findings highlight reflection as a powerful and underexplored inductive bias for scalable, generalizable robot learning.

\section{Limitations}

\subsection{Applicability to Other Camera Placements}

\textit{Eye-in-hand-only} setups remove visual asymmetry from the third-person camera, and do not require centering in image space, allowing direct application without modification. The mirrored setup then corresponds to the initial camera frame.

As discussed in Sec.~\ref{sec:methodology}, \textit{off-centered third-person cameras} can be addressed by pre-centering the end-effector tool-center point (TCP). 
The TCP image coordinate can be obtained either through hand–eye calibration and robot kinematics, or by leveraging vision models such as Grounded-SAM~\cite{ren2024grounded} (App.~F).

\textit{Side-camera} setups typically have the end-effector near the image center, and as shown in App. F, centering on the end-effector makes MirrorDuo remain effective. 
As explained in the above section, asymmetries in the background and robot, which may be more pronounced in side-camera settings, can be narrowed using visual generalization and a few in-domain samples. 
However, unlike front-camera setups, side cameras may imply a more offset initial configuration, potentially leading to mirrored poses outside the workspace.

While the formulation remains unchanged, \textit{over-the-shoulder} cameras can introduce more self-occlusion, potentially widening the visual gap and warranting further testing. Interestingly, in symmetric bimanual setups with dual-arm robots and over-the-shoulder cameras mimicking head position, we expect mirrored trajectories to transfer directly to the other arm.

\subsection{Marginal Performance Drop under Sufficient Data}

\begin{figure}[h]
        \centering
        \includegraphics[width=0.85\linewidth]{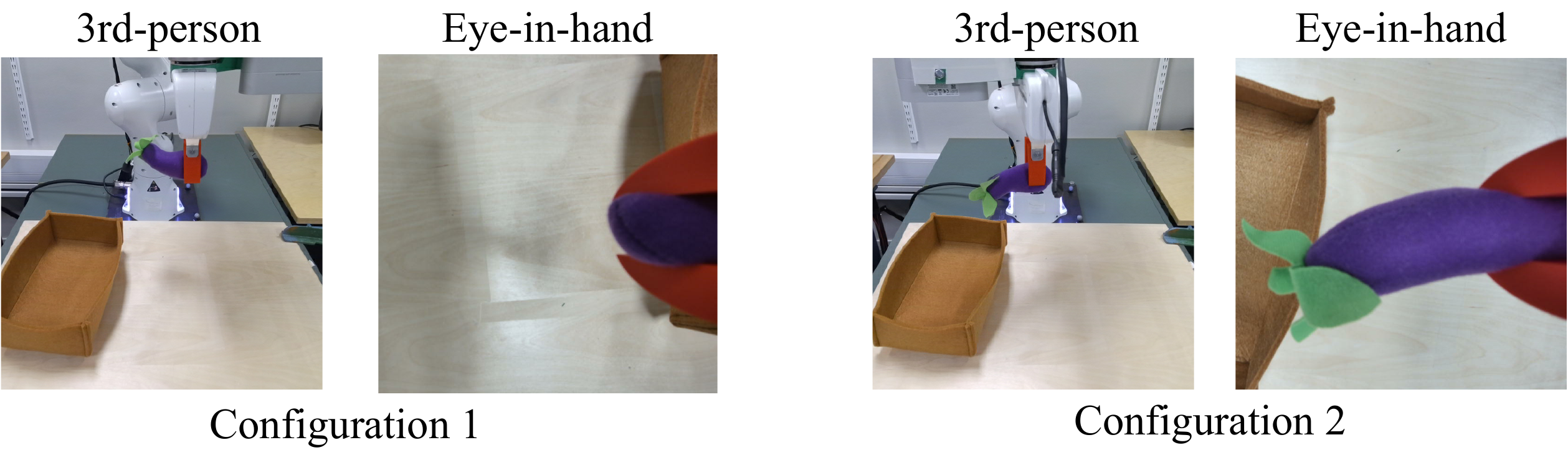}
        \captionof{figure}{
        Illustration of two plausible robot configurations and eye-in-hand views that share near-identical third-person observations.
        }
        \label{fig:decision_boundary}
        \vspace{-2mm}
    \end{figure}

As shown in Tab.~\ref{tab:main_results}, both MirrorDiffusion and Diffusion + MirrorAug exhibit a slight performance drop on \textit{Square D2} and \textit{Three-piece Assembly D1} compared to their baseline counterparts. 
We hypothesize this is due to fragmented decision boundaries arising when near-identical task setups yield differing observation distributions. Specifically, the eye-in-hand camera provides only a portion of the workspace. 
As illustrated above, depending on the task, the demonstrator may choose either a clockwise or counter-clockwise rotation, leading to two distinct eye-in-hand perspectives and action trajectories. 
While mirrored trajectories maintain semantic consistency with the source trajectory, they create inconsistent mappings across the mirrored pairs. 
This results in additional, possibly conflicting, decision boundaries that may degrade performance. See App.~E for more qualitative examples.
However, in low- to medium-data regimes where demonstrations are sparse, this additional diversity can be beneficial by enriching the training distribution and helping the policy generalize.


\clearpage
\acknowledgments{This work was partially supported by the Wallenberg AI, Autonomous Systems and Software Program (WASP) funded by the Knut and Alice Wallenberg Foundation.}


\bibliography{references}  

\clearpage


\section*{\textit{A}\quad Formulation Derivations}
\subsection*{Eye-in-hand Local-frame Reparameterization}

For an eye-in-hand camera setup, let the current camera frame at time \( t \) be denoted as \( \mathbf{X}_{C_t} \in \mathrm{SE}(3) \), where \( \mathrm{SE}(3) \) represents the Special Euclidean group. 
The initial and previous camera frames are denoted as \( \mathbf{X}_{C_0} \) and \( \mathbf{X}_{C_{t-1}} \), respectively.
Let the end-effector pose at time \( t \) be denoted as \( \mathbf{X}_{H_t} \). 
Expressed in the initial eye-in-hand camera frame, and relative to the initial end-effector pose, the delta transformation \( \Delta \mathbf{X}_{H_t} \) is:
\begin{equation*}
    \Delta \mathbf{X}_{H_t}^{\phantom{*}} 
    := 
   \mathbf{X}_{H_0}^{-1} \mathbf{X}_{H_t}^{\phantom{*}} = \left({}^{C_0}\mathbf{X}_{H_0}^{\phantom{*}}\right)^{-1} 
   \left({}^{C_0}\mathbf{X}_{H_t}^{\phantom{*}}\right).
\end{equation*}
Similarly, the relative pose with respect to the previous timestep is:
\begin{equation*}
    \delta \mathbf{X}_{H_t}^{\phantom{*}}
    := \mathbf{X}_{H_{t-1}}^{-1}\mathbf{X}_{H_t}^{\phantom{*}}
    = \left({}^{C_{t-1}}\mathbf{X}_{H_{t-1}}^{\phantom{*}}\right)^{-1} \left({}^{C_{t-1}}\mathbf{X}_{H_t}^{\phantom{*}}\right).
\end{equation*}

\subsection*{Fixed Points of the Pose Mirroring Mapping}

For successful transfer of the mirrored trajectory to the current robot configuration, it is crucial that the initial state of the manipulation trajectory lies near a fixed point of the pose mirroring mapping.
That is, the pose should remain close to configurations that are invariant under the mirror mapping \( \mathcal{M}(\cdot) \). For an arbitrary pose \( \mathbf{X} \in \mathrm{SE}(3) \), the mirroring mapping is defined as \( \mathcal{M}(\mathbf{X}) = \mathbf{E} \mathbf{X} \mathbf{E} \), as introduced in Eq.~\eqref{eq:abs_pose_mirroring}, where \( \mathbf{E} = \mathrm{diag}([-1, 1, 1, 1]) \).
The fixed points of this mapping must satisfy $
\mathbf{E} \mathbf{X} \mathbf{E} = \mathbf{X}$.
Applying the mirroring operation to a general pose \( \mathbf{X} \in \mathrm{SE}(3) \):
\begin{equation*}
\mathbf{E}
\begin{bmatrix}
r_{xx} & r_{yx} & r_{zx} & t_x \\
r_{xy} & r_{yy} & r_{zy} & t_y \\
r_{xz} & r_{yz} & r_{zz} & t_z \\
0      & 0      & 0      & 1
\end{bmatrix}
\mathbf{E}
=
\begin{bmatrix}
r_{xx} & -r_{yx} & -r_{zx} & -t_x \\
-r_{xy} & r_{yy} & r_{zy} & t_y \\
-r_{xz} & r_{yz} & r_{zz} & t_z \\
0       & 0      & 0      & 1
\end{bmatrix}.
\end{equation*}
Therefore, for a pose to be a fixed point under \( \mathcal{M} \), the following symmetry conditions must hold:
\[
r_{yx} = r_{zx} = r_{xy} = r_{xz} = 0, \quad t_x = 0,
\]
which means that the rotation matrix corresponds to a pure rotation about the x-axis.
The translation vector is only mildly affected by the local reparameterization ($\delta \mathbf{X}_{H_t}$, $\Delta \mathbf{X}_{H_t}$) around the origin (i.e., \( \mathbf{t}^* \approx \mathbf{t} \) for small motions). Likewise, small rotational motions that follow this fixed-point structure, i.e., rotations about the x-axis, are only slightly perturbed by the mirroring operation.

\section*{\textit{B}\quad Reflection-Equivariant 
Diffusion Policy (MirrorDiffusion)}

\begin{figure}[t]
    \centering
    \includegraphics[width=0.95\linewidth]{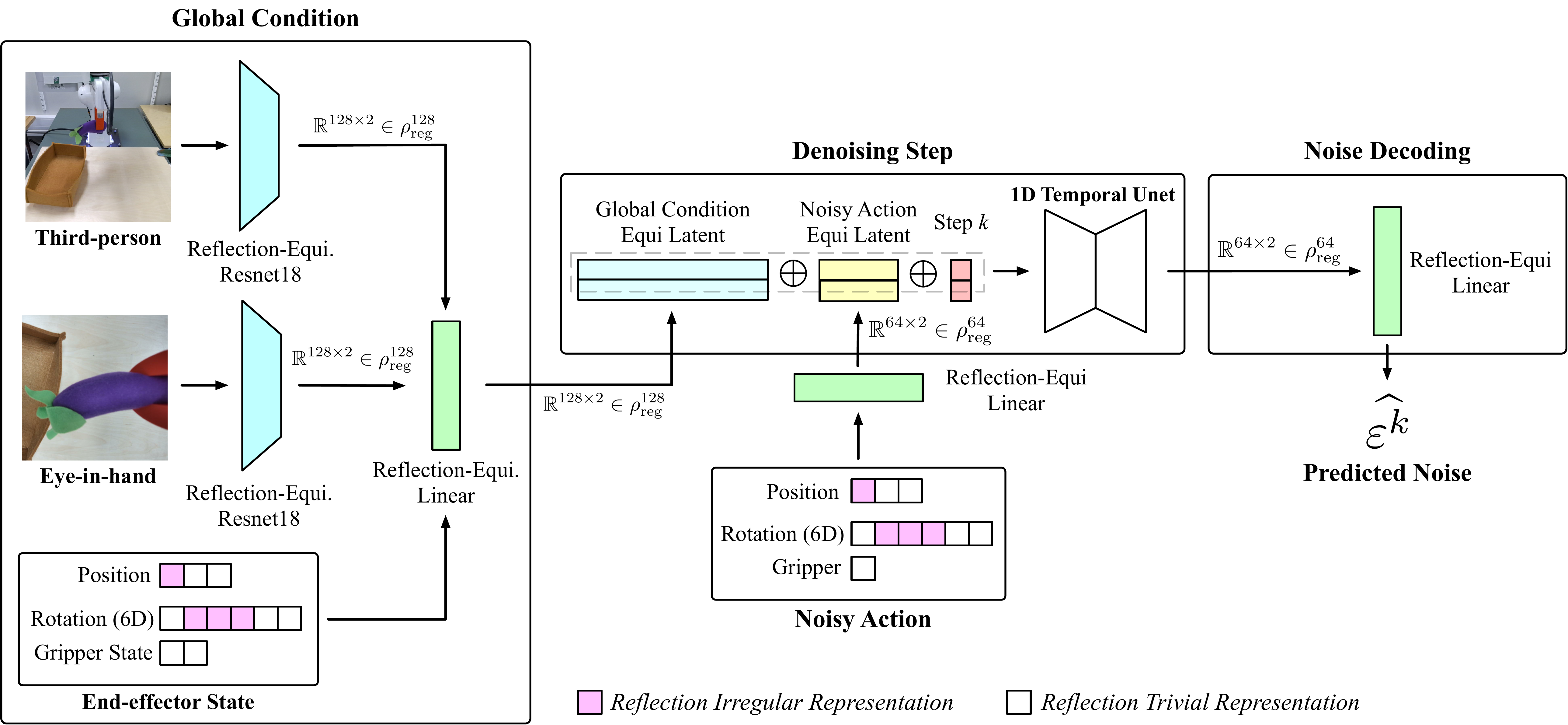}
    \caption{Illustration of Reflection Equivariant Diffusion (MirrorDiffusion) Network Architecture.
    }
    \label{fig:mirrordiff_net}
    \vspace{-2mm}
\end{figure}

The general architecture of MirrorDiffusion follows the $\mathrm{SO}(2)$-Equivariant Diffusion Policy proposed by~\citet{wang_equivariant_diff_po}, with the key difference being a change in structural equivariance from rotation to reflection. 
As illustrated in Fig.~\ref{fig:mirrordiff_net}, the Equivariant ResNet used in~\cite{wang_equivariant_diff_po} is modified to be reflection-equivariant by constructing the ResNet architecture using the abstract group \texttt{Flip2dOnR2} provided in the E(n)-CNN library~\cite{cesa2022_En_CNN}. 
The end-effector states are arranged following the representation specified by the color coding in the figure. The reflection-equivariant linear layers are implemented by overloading the Dihedral group in the E(n)-CNN library~\cite{cesa2022_En_CNN}, with the number of group elements set to 1 (a group only contains the original element and reflected counter part).

During the \textit{encoding phase} (generating the global condition), two independent reflection-equivariant ResNets encode the third-person and eye-in-hand views, each producing a pair of 128-dimensional regular representations (i.e.,  \(128 \times 2\)).
The robot states are arranged according to the corresponding irregular and trivial group representations, as formulated in Eq.~\eqref{eq:mirror_vec_form} and illustrated in Fig.~\ref{fig:mirrordiff_net}. 
A subsequent reflection-equivariant linear layer encodes these mixed representations into a \(128 \times 2\) regular representation.
During the \textit{denoising phase}, the noisy action is arranged according to the representation specified in Fig.~\ref{fig:mirrordiff_net}, and is processed by a reflection-equivariant linear layer, producing a \(64 \times 2\) regular representation. 
A 1D Temporal U-Net with hidden dimensions \([512, 1024, 2048]\) then processes each element of the embedding. Conceptually, this corresponds to applying the U-Net independently to the concatenated embedding (comprising the global condition and the action embedding) for both the original and reflected inputs, yielding a 64-dimensional embedding for each component.
The separately denoised latents are recovered to shape \(64 \times 2\), which is then passed through a final reflection-equivariant decoder to produce the predicted noise.

\section*{\textit{C}\quad Experiment Implementation Details}
\subsubsection*{Baseline Networks}
\textit{Diffusion Policy} follows the hybrid-CNN architecture with global conditioning as described in~\cite{chi2023diffusionpolicy}, consistent with the baseline implementation in~\cite{wang_equivariant_diff_po}.
The input horizon, action horizon, and action prediction horizon are set to 2, 8, and 16, respectively. 
A fixed learning rate of \(1 \times 10^{-4}\) is used. The model utilizes a DDPM noise scheduler~\cite{ho2020denoising}, with both training and inference configured for 100 diffusion steps across simulated and real-world experiments.

\textit{BC-RNN} follows the network architecture and hyperparameters specified in RoboMimic~\cite{robomimic2021}. Specifically, the image encoder comprises a ResNet18~\cite{he2016deep} followed by a Spatial Softmax layer The extracted image features are concatenated with the robot's proprioceptive states and passed through a 2-layer LSTM, whose final hidden state is used as input to a Gaussian Mixture Model (GMM) policy head. As in RoboMimic~\cite{robomimic2021}, during rollout, the learned standard deviations of each GMM mode are clamped and replaced with a fixed value of \(1 \times 10^{-4}\).

\subsubsection*{Random Overlay}
Random Overlay plays an integral role in MirrorDuo.  
For diffusion policies, we follow the default setting described in~\cite{zhuang2024robosaga}.
Specifically, for each batch of trajectories, we randomly sample half and overlay their images with random backgrounds using a blend factor \(\alpha = 0.5\). 
During preliminary experiments, MirrorDiffusion exhibited minor performance degradation under stronger overlays (i.e., lower \(\alpha\)). 
As a result, we set the blend factor to \(\alpha = 0.75\). 
The blending operation is defined as:
\[
\texttt{overlaid\_image} = \alpha \cdot \texttt{image} + (1 - \alpha) \cdot \texttt{random\_background}, \quad \alpha \in [0, 1].
\]
The number of warmup epochs, during which the ratio of sampled trajectories gradually ramps up to the designated threshold, is set to $\min(20,\ 4000 / \texttt{num\_demos})$ to accommodate varying numbers of demonstrations.

\subsubsection*{Training Epochs and Evaluation Protocols}
The total number of training epochs is scaled according to the number of demonstrations, computed as \(50000 / \texttt{num\_demos}\). Evaluation is performed every \(2000 / \texttt{num\_demos}\) steps. 
At each evaluation step, 50 rollouts are performed.
For experiments involving additional demonstrations from mirrored arrangements, the number of demonstrations used to compute the total training epochs and evaluation frequency is fixed to the base number of demonstrations, i.e., 200.

For the data points of the $\mathrm{SO}(2)$-Equivariant Diffusion Policy~\cite{wang_equivariant_diff_po} in Table~\ref{tab:main_results} and Table~\ref{tab:appen_full_table}, results for tasks with 100 and 200 demonstrations are directly taken from the published results. Results for 50 and 500 demonstrations are not publicly available and are therefore newly generated using the authors' released code.

\textbf{Image size}. The image inputs for all experiments are of size \(3 \times 84 \times 84\), with a random crop of size \(76 \times 76\) applied during training.
The crop is set to \(76 \times 76\) center crop during evaluation.

\textbf{Initial Pose.} The local reparameterization of poses and actions (Eq.~\eqref{eq:mirror_relative}) require centering all trajectories around a fixed initial pose.  
In simulation, where the starting pose is constant, we use this fixed pose directly.  
In real-world experiments, where initial poses vary within a neighborhood, we use the average initial pose across demonstrations.

\subsubsection*{Simulation Task Descriptions}

In this work, we use five simulation tasks from MimicGen~\cite{mandlekar2023mimicgen}, using the provided datasets.
All tasks employ the Franka Panda robot as the manipulator, operating in a 7-dimensional action space comprising 6 degrees of freedom for the end-effector pose and 1 dimension for gripper open/close.
Each task uses two camera views: a third-person view and an eye-in-hand view. 
Task descriptions and key properties are summarized below:

\begin{itemize}
\item \textit{Square D0}: Grasp the square nut by the handle and insert it into a matching square peg. The nut undergoes 360° random rotation around the z-axis, with limited positional variation. The target peg remains fixed.
\item \textit{Square D2}: Same objective as Square D0, but with a broader distribution over both the nut’s and peg’s positions and orientations.
\item \textit{Coffee D2}: Pick up the coffee pod from one side, insert it into the coffee machine on the opposite side, and close the lid. The coffee pod has constrained positional variation, and the coffee machine has limited variation in position and z-axis orientation.
\item \textit{Stack Three D1}: Sequentially stack three cubes on top of each other. Positions and z-axis orientations of the cubes are randomized within the workspace.
\item \textit{Three Piece Assembly D1}: Sequentially assemble three pieces, requiring stricter precision on orientation and placement.
\end{itemize}

Except for \textit{Square D0} and \textit{Coffee D2}, which involve constrained initialization, all other tasks allow full 360° rotation around the z-axis and broad position variation for all relevant objects.


\section*{\textit{D}\quad Simulation Results with Standard Deviation}
Table~\ref{tab:appen_full_table} presents the complete simulation results from Table~\ref{tab:main_results}, including standard deviations.

\begin{table}[h]
    \centering
    \scriptsize
    \definecolor{gainblue}{RGB}{58,147,195}
    \definecolor{gainred}{RGB}{215,95,76}
    \setlength{\tabcolsep}{4pt}
    \begin{tabular}{l@{\hskip 4pt}lccc ccc ccc}
    \toprule
    & & \multicolumn{3}{c}{\textbf{Stack Three (D1)}} & \multicolumn{3}{c}{\textbf{Square (D2)}} & \multicolumn{3}{c}{\textbf{3-Part Assembly (D2)}} \\
    \cmidrule(lr){3-5} \cmidrule(lr){6-8} \cmidrule(lr){9-11}
    & \textbf{Method} & \textit{50} & \textit{100} & \textit{200} & \textit{100} & \textit{200} & \textit{500} & \textit{100} & \textit{200} & \textit{500} \\
    \midrule
    \multirow{6}{*}{\rotatebox[origin=c]{90}{\textbf{Delta}}}
    & EquiDiff. & 20.7$\pm$0.9 & 54.7$\pm$5.2 & 77.3$\pm$1.8 & 25.3$\pm$8.7 & 41.3$\pm$9.8 & 60.0$\pm$7.5 & 15.3$\pm$1.8 & 39.3$\pm$1.8 & 63.0$\pm$3.0\\
    & MirrorDiff. & 51.3$\pm$0.9 & 77.3$\pm$0.9 & 89.3$\pm$0.9 & 24.0$\pm$2.8 & 48.7$\pm$2.5 & 59.3$\pm$3.4 & 25.3$\pm$2.5 & 49.3$\pm$3.4 & 61.3$\pm$4.1\\
    \cmidrule(){2-11}
    & DiffPo. & 19.3$\pm$3.8 & 47.3$\pm$2.5 & 80.7$\pm$5.0 & 20.7$\pm$0.9 & 40.0$\pm$2.8 & 58.0$\pm$1.6 & 11.0$\pm$3.0 & 31.3$\pm$2.5 & 64.0$\pm$7.5\\
    & DiffPo. + $\mathcal{M}$ & 50.7$\pm$1.9 & 68.0$\pm$3.3 & 91.3$\pm$0.9 & 32.7$\pm$2.5 & 49.3$\pm$8.4 & 56.7$\pm$2.5 & 21.0$\pm$1.0 & 47.3$\pm$6.6 & 61.3$\pm$5.0\\
    \cmidrule(){2-11}
    & BC-RNN & 0.7$\pm$0.9 & 2.0$\pm$0.0 & 8.0$\pm$1.6 & 4.0$\pm$1.6 & 9.3$\pm$2.5 & 32.0$\pm$4.3 & 0.0$\pm$0.0 & 0.0$\pm$0.0 & 6.0$\pm$0.0\\
    & BC-RNN + $\mathcal{M}$ & 0.0$\pm$0.0 & 3.3$\pm$1.9 & 37.3$\pm$6.2 & 4.7$\pm$2.5 & 12.7$\pm$0.9 & 43.3$\pm$8.2 & 1.0$\pm$1.0 & 3.3$\pm$1.9 & 12.0$\pm$2.8\\
    \midrule
    \multirow{6}{*}{\rotatebox[origin=c]{90}{\textbf{Relative}}}
    & EquiDiff. & 6.7$\pm$2.5 & 25.3$\pm$3.3 & 62.7$\pm$3.5 & 11.3$\pm$1.3 & 20.7$\pm$4.1 & 40.0$\pm$2.0 & 1.3$\pm$0.7 & 4.7$\pm$0.7 & 22.0$\pm$2.8\\
    & MirrorDiff. & 28.7$\pm$2.5 & 57.3$\pm$0.9 & 80.0$\pm$3.3 & 18.0$\pm$1.6 & 32.0$\pm$3.3 & 47.3$\pm$1.9 & 13.3$\pm$2.5 & 23.3$\pm$2.5 & 50.0$\pm$1.6\\
    \cmidrule(){2-11}
    & DiffPo. & 19.3$\pm$0.9 & 31.3$\pm$3.4 & 58.0$\pm$3.3 & 18.0$\pm$3.3 & 30.0$\pm$4.3 & 44.0$\pm$1.6 & 4.0$\pm$0.0 & 13.3$\pm$0.9 & 32.0$\pm$4.3\\
    & DiffPo. + $\mathcal{M}$ & 29.3$\pm$3.4 & 50.0$\pm$2.8 & 68.0$\pm$0.0 & 32.7$\pm$2.5 & 41.3$\pm$3.4 & 45.3$\pm$5.7 & 11.0$\pm$1.0 & 26.7$\pm$5.7 & 48.0$\pm$1.6\\
    \cmidrule(){2-11}
    & BC-RNN & 6.7$\pm$4.1 & 18.0$\pm$1.6 & 51.3$\pm$11.1 & 8.0$\pm$1.6 & 19.3$\pm$2.5 & 45.3$\pm$3.8 & 2.0$\pm$0.0 & 3.3$\pm$1.9 & 11.3$\pm$5.7\\
    & BC-RNN + $\mathcal{M}$ & 18.0$\pm$5.7 & 35.3$\pm$6.8 & 73.3$\pm$3.4 & 16.0$\pm$2.8 & 24.7$\pm$0.9 & 48.0$\pm$8.6 & 1.0$\pm$1.0 & 8.7$\pm$2.5 & 22.7$\pm$1.9\\
    \bottomrule
    \end{tabular}
    \vspace{1mm}
    \caption{\textbf{Setting~III: Wide-view, two-sided demonstrations.} Success rate (\%) on three MimicGen tasks as the number of demonstrations increases. \textit{Delta} and \textit{Relative} refer to action controllers. Results are averaged over the top-1 rollout (50 trials) from three training seeds. EquiDiff\. denotes the SO(2)-equivariant diffusion policy~\cite{wang_equivariant_diff_po}, DiffPo. denotes the diffusion policy~\cite{chi2023diffusionpolicy}, $\mathcal{M}$ denotes MirrorDuo augmentation, and MirrorDiff. denotes the proposed mirror-equivariant diffusion policy.}
    \label{tab:appen_full_table}
    \vspace{-2mm}
\end{table}

\section*{\textit{E}\quad Mismatch Between Mirrored and Actual Demonstrations}

\begin{figure}[t]
  \centering
  \subfloat[Original Demo A \label{fig:demo_a}]{
    \includegraphics[width=0.35\linewidth]{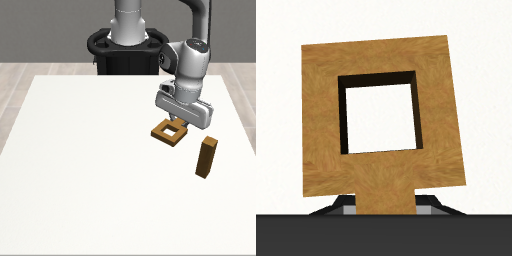}
  }
    \hspace{5mm}
  \subfloat[Mirrored Demo A\label{fig:demo_a_mirrored}]{
    \includegraphics[width=0.35\linewidth]{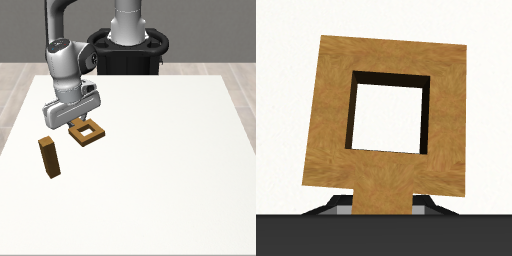}
  }
  \\
  \subfloat[Original Demo B\label{fig:demo_b}]{
    \includegraphics[width=0.35\linewidth]{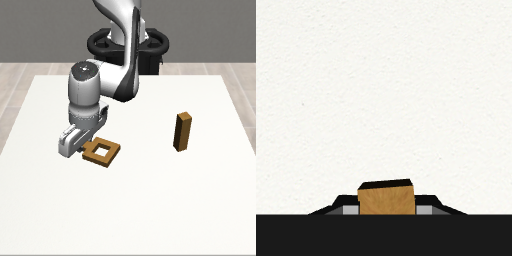}
  }
    \hspace{5mm}
  \subfloat[Mirrored Demo B]{
    \includegraphics[width=0.35\linewidth]{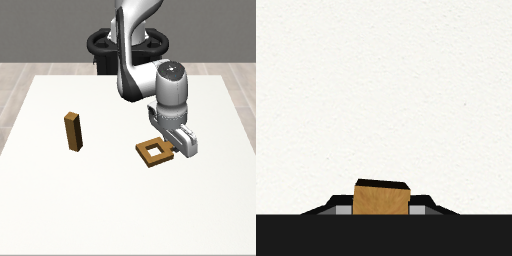}
  }
    \caption{
    \textbf{Illustration of Conflicting Visual Cues and Trajectories} introduced by mirrored demonstrations in the \textit{Square D2} task. 
      Each mirrored demonstration features an approximately co-located square nut relative to its original counterpart (e.g., Fig.(b, c) and Fig.(d, a)), yet exhibits a distinct eye-in-hand view.
    This discrepancy suggests that while the mirrored and original demonstrations share a similar initial setup (i.e., the first subtask), they require oppositely rotating actions.
    }
\label{fig:migalign_square_d2}
\vspace{-2mm}
\end{figure}

\begin{figure}[t]
  \centering
  \subfloat[Original Demo A \label{fig:assem_demo_a}]{
    \includegraphics[width=0.35\linewidth]{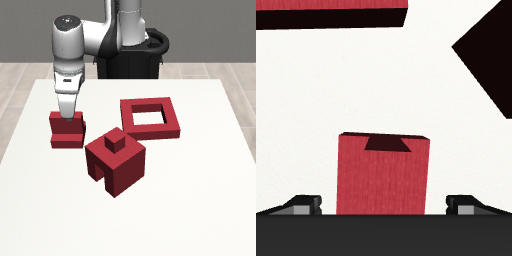}
  }
    \hspace{5mm}
  \subfloat[Mirrored Demo A\label{fig:assem_demo_a_mirrored}]{
    \includegraphics[width=0.35\linewidth]{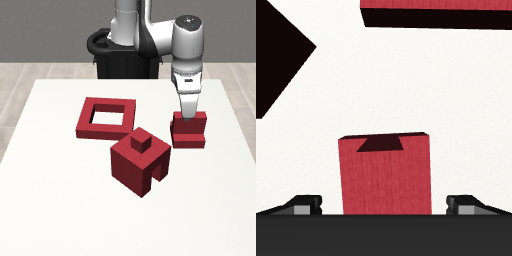}
  }
  \\
  \subfloat[Original Demo B\label{fig:assem_demo_b}]{
    \includegraphics[width=0.35\linewidth]{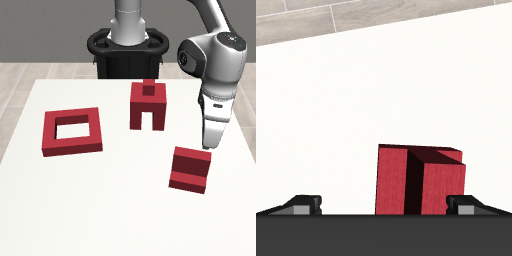}
  }
    \hspace{5mm}
  \subfloat[Mirrored Demo B\label{fig:assem_mirrored_demo_b}]{
    \includegraphics[width=0.35\linewidth]{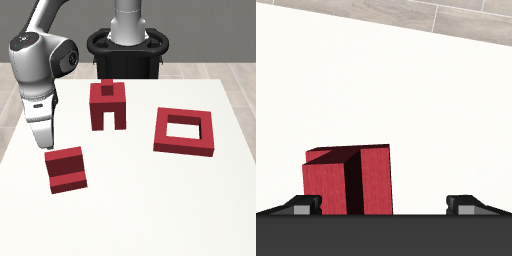}
  }
\caption{
    \textbf{Illustration of Conflicting Visual Cues and Trajectories} introduced by mirrored demonstrations in the \textit{Three Piece Assembly D1} task. 
    Each mirrored demonstration features an approximately co-located T-shaped piece relative to its original counterpart (e.g., (b, c) and (d, a)), yet exhibits a distinct eye-in-hand view, one oriented toward the workspace, the other facing outward. 
    This discrepancy suggests that while the mirrored and original demonstrations share a similar initial setup (i.e., the first subtask), they require oppositely rotating actions.
}
\label{fig:migalign_assembly}
\vspace{-2mm}
\end{figure}

As discussed in the limitations section and observed in Table~\ref{tab:main_results}, when the number of demonstrations increases to 500 for the \textit{Square D2} and \textit{Three Pieces Assembly D1} tasks, MirrorDuo exhibits a marginal decrease in performance. 
We hypothesize that this is due to the mirrored demonstrations increasing the level of multi-modality in the data, leading to a more fragmented decision boundary.

Specifically, for a given original demonstration, its mirrored counterpart may represent a valid but conflicting trajectory from the actual sample contained in the original dataset.
For \textit{Square D2}, as illustrated in Fig.~\ref{fig:migalign_square_d2}, the mirrored demonstration in Fig.~\ref{fig:demo_a_mirrored} shows the square nut approximately co-located with that in the other original demonstration (Fig.~\ref{fig:demo_b}). However, the mirrored eye-in-hand view (Fig.~\ref{fig:demo_a_mirrored}) shows the entire square nut clearly, while in the original view, only the handle of the nut is visible. 
Fig.~\ref{fig:migalign_assembly} shows one of the examples in the \textit{Three-piece Assembly D1} Task.
Each mirrored demonstration features an approximately co-located T-shaped piece relative to its original counterpart (e.g., Fig. (b, c) and Fig. (d, a)), yet exhibits a distinct eye-in-hand view, one oriented toward the workspace, the other facing outward. 

These examples indicate that, under similar subtask setups, the augmented data introduces additional valid trajectories that involve opposing end-effector rotations and result in distinct eye-in-hand views. This divergence introduces ambiguity into the learned policy, and the likelihood of such ambiguity increases as the density of demonstrations grows.
\section*{\textit{F}\quad MirrorDuo with Off-Centered Third-Person Camera}
\label{app:off_center}
\begin{figure}[h]
  \centering
  \subfloat[Original Image]{
    \includegraphics[width=0.23\linewidth]{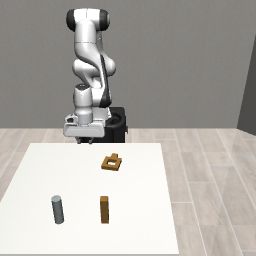}
  }
  \hfill
  \subfloat[Roll-Centered Image]{
    \includegraphics[width=0.23\linewidth]{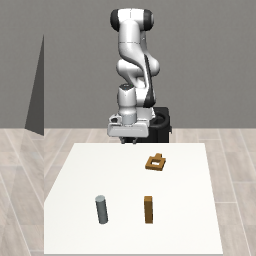}
  }
  \hfill
  \subfloat[Centered Mirror Image]{
    \includegraphics[width=0.23\linewidth]{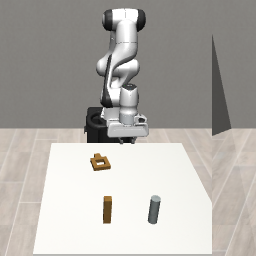}
  }
    \hfill
  \subfloat[Centered Mirror Setup]{
    \includegraphics[width=0.23\linewidth]{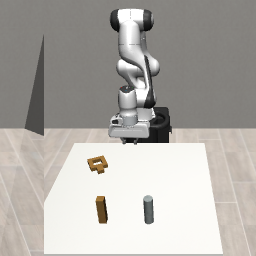}
  }
\caption{
    \textbf{Illustration of Mirroring with an Off-Centered Camera.}
    (a) Original image from an off-centered third-person camera. 
    (b) Roll-centered image with the end-effector aligned to the mirroring axis. 
    (c) Mirrored version of the roll-centered image used by MirrorDuo. 
    (d) Roll-centered image from the actual mirrored setup.
}
\label{fig:off-centered}
\vspace{-3mm}
\end{figure}

\begin{table}[t]
    \centering
    \begin{tabular}{lcccc}
        \toprule
        & \multirow{2}{*}{\textbf{In-domain}} & \multicolumn{3}{c}{\textbf{\# M-Demos}} \\
        \cmidrule(lr){3-5}
        & & 0 & 5 & 10 \\
        \midrule
        MirrorDiff. & \textbf{89.3} $\pm$ 1.9 & 0.0 $\pm$ 0.0 & \textbf{72.7} $\pm$ 1.9 & \textbf{90.0} $\pm$ 1.6 \\
        DiffPo. + $\mathcal{M}$ & 83.3 $\pm$ 1.9 & 0.0 $\pm$ 0.0 & 69.3 $\pm$ 0.9 & 84.0 $\pm$ 1.6 \\
        DiffPo. & 85.3 $\pm$ 2.5 & 0.0 $\pm$ 0.0 & 23.3 $\pm$ 1.9 & 32.7 $\pm$ 3.8 \\
        \bottomrule
    \end{tabular}
    \vspace{1mm}
    \caption{
    \textbf{Off-Centered Third-Person Camera, One-Sided.} Success rate (\%) on the \textit{Square D0} task under the \textit{mirrored arrangement}, with an additional 5 or 10 demonstrations from the mirrored setup (denoted as M-Demos in this table) added on top of the original 200 demonstrations. 
    Each data point reports the average of the top-3 evaluations, with 50 rollouts per evaluation.
    }
    \label{tab:off_centered_square_d0}
    \vspace{-2mm}
\end{table}

\begin{figure}[t]
    \centering
    \begin{minipage}[b]{0.58\linewidth}
   \centering
    \begin{tabular}{lcc}
        \toprule
        & \multicolumn{2}{c}{\textbf{Third-person Camera}} \\
        \cmidrule(lr){2-3}
        & Centered  & Off-centered \\
        \midrule
        MirrorDiff. &  89.3 $\pm$ 0.9 &  \textbf{84.7} $\pm$ 1.9 \\
        DiffPo. + $\mathcal{M}$ & \textbf{91.3} $\pm$ 0.9 & 81.3 $\pm$ 1.9 \\
        DiffPo. & 80.7 $\pm$ 5.0 & 70.7 $\pm$ 3.4 \\
        \bottomrule
    \end{tabular}
    \vspace{1mm}
    \captionof{table}{
    \textbf{Off-Centered Third-Person Camera, Two-Sided.} Success rates (\%) on the \textit{Stack Three D1} task.  
    }
    \label{tab:off_center_stack_three}
    \end{minipage}
    \hfill
    \begin{minipage}[b]{0.38\linewidth}
    \centering
    \includegraphics[width=0.5\linewidth]{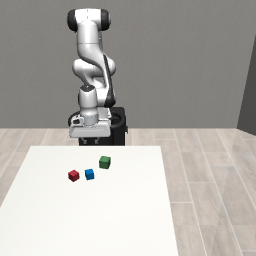}
    \captionof{figure}{Illustration of Off-centered camera view for \textit{Stack Three D1}
    }
    \label{fig:off_center_stack_three}
    \end{minipage}
    \vspace{-2mm}
\end{figure}

To transfer the mirrored skill to the initial configuration based on the given demonstrations, MirrorDuo requires alignment not only in the proprioceptive states but also in the image space. 
This means that the end-effector should be near the mirroring axis, i.e. the horizontal center of the image.
Although random cropping alleviates the strictness of centering to the midline, a general alignment is still required. For off-centered third-person cameras, a pre-alignment step is necessary.
Otherwise, even if MirrorDuo successfully learns the mirrored skill, the starting configuration and workspace setup of the mirrored demonstration will not align with the ideal scenario of transferring under the current robot configuration to a mirrored object arrangement.

Here we show that MirrorDuo can be applied to off-centered third-person camera scenarios by pre-centering the view, demonstrated in two settings: one with one-sided demonstrations (\textit{Square D0}, Fig.~\ref{fig:off-centered}) and another with two-sided demonstrations (\textit{Stack Three D1}, Fig.~\ref{fig:off_center_stack_three}).
Let the off-centered camera be denoted as \{C\} and the re-centered camera as \{$C_{\mathrm{ref}}$\}. The mirrored setup is derived using Eq.~\eqref{eq:abs_pose_mirroring}, where the mirroring is applied with respect to the re-centered camera pose, i.e., $\mathbf{X}_{C_{\mathrm{ref}}}$.

Following previous setups, we evaluate MirrorDiffusion, Diffusion + MirrorAug, and the Diffusion baseline using re-rendered demonstrations under off-centered cameras. 
In the one-sided case, we assess performance on mirrored arrangements with 0, 5, and 10 additional demonstrations. 
For the two-sided case, we directly evaluate in-domain performance. 
All experiments assume access to global camera extrinsics, enabling roll-centering by aligning the initial end-effector pose to the image center. 
The same offset is applied to subsequent frames, with the rolled region tinted green to indicate shifted areas. 
Networks receive these centered images as input.

In the off-centered camera settings, the visual domain gap between the (already-centered) mirrored and original samples arises not only from the robot’s asymmetry but also from perspective shifts across the left and right sides of the workspace. For instance, as shown in Fig.~\ref{fig:off-centered}, in the original domain, the square peg appears near the horizontal center of the image, showing only its front face. In the mirrored setup, however, the peg shifts toward the left side of the image, revealing its right face, an angle not observed in the original demonstrations. Additionally, the appearance of the table also changes under this off-centered view.

Table~\ref{tab:off_centered_square_d0} shows that the widened visual domain gap reduces the performance of direct transfer to the mirrored arrangement to zero, in contrast to the matched in-domain performance of Diffusion + MirrorAug when the third-person camera is centered (Fig.~\ref{fig:visual_asymm_from_robot}).
However, the data efficiency benefit of MirrorDuo remains evident. 
Under the increased visual domain gap, the performance in the mirrored arrangement with five additional demonstrations is only 17\% and 24\% lower than the corresponding in-domain setting for MirrorDiffusion and Diffusion + MirrorAug, respectively. 
With ten additional demonstrations, both methods recover their performance in the mirrored setup, matching their in-domain success rates. In contrast, without MirrorDuo, simply adding ten demonstrations in the mirrored setup results in only a 32.7\% success rate for the baseline diffusion policy.

For the two-sided setup with an off-centered camera, we evaluate on \textit{Stack Three D1}. The centered camera results reported in Table~\ref{tab:off_center_stack_three}, are drawn from Table~\ref{tab:appen_full_table} (averaged over three seeds). For the off-centered case, each entry reflects the average of the top 3 evaluations, with 50 rollouts per evaluation due to limited compute resources. As shown in Table~\ref{tab:off_center_stack_three}, all methods exhibit a performance drop under the more challenging viewpoint. MirrorDiffusion and Diffusion Policy + MirrorAug maintain relatively high success rates (84.7\% and 81.3\%, respectively), while the baseline drops to 70.7\%. This $\sim$10\% decline aligns with trends observed in the centered-camera setting, highlighting the effectiveness of mirroring-based approaches to off-centered camera variations.

\end{document}